\documentclass[sigconf]{acmart}
\usepackage{microtype}
\usepackage{graphicx}
\usepackage{booktabs}
\usepackage{colortbl}
\usepackage{multicol}
\usepackage{multirow}
\usepackage{float}
\usepackage{footmisc}
\usepackage{url}
\usepackage{makecell}
\usepackage[ruled,vlined]{algorithm2e}
\usepackage{algpseudocode}
\usepackage{subfigure}
\usepackage{mathtools}
\usepackage{nicematrix}
\usepackage{booktabs}
\usepackage{caption}

\definecolor{'wit'}{HTML}{FBFBFB}
\definecolor{'gry'}{HTML}{EEEEEE}

\definecolor{'deep1'}{HTML}{C5E6F8} 
\definecolor{'shallow1'}{HTML}{E4F3FC} 
\definecolor{'deep2'}{HTML}{E5F5B7} 
\definecolor{'shallow2'}{HTML}{F3FADF} 

\definecolor{'deep3'}{HTML}{FFE5C6} 
\definecolor{'shallow3'}{HTML}{FFF2E3} 
\definecolor{'deep4'}{HTML}{FFD3CF} 
\definecolor{'shallow4'}{HTML}{FFEAE8}
\definecolor{'deep5'}{HTML}{D2D0F3} 
\definecolor{'shallow5'}{HTML}{E8E7F9} 
\algrenewcommand\algorithmicrequire{\textbf{Input:}}

\copyrightyear{2024}
\acmYear{2024}
\setcopyright{rightsretained}
\acmConference[KDD '24]{Proceedings of the 30th ACM SIGKDD Conference on Knowledge Discovery and Data Mining}{August 25--29, 2024}{Barcelona, Spain}
\acmBooktitle{Proceedings of the 30th ACM SIGKDD Conference on Knowledge Discovery and Data Mining (KDD '24), August 25--29, 2024, Barcelona, Spain}\acmDOI{10.1145/3637528.3671564}
\acmISBN{979-8-4007-0490-1/24/08}
\makeatletter
\gdef\@copyrightpermission{
  \begin{minipage}{0.3\columnwidth}
   \href{https://creativecommons.org/licenses/by-nc-sa/4.0/}{\includegraphics[width=0.90\textwidth]{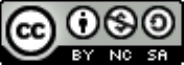}}
  \end{minipage}\hfill
  \begin{minipage}{0.7\columnwidth}
   \href{https://creativecommons.org/licenses/by-nc-sa/4.0/}{This work is licensed under a Creative Commons Attribution-NonCommercial-ShareAlike International 4.0 License.}
  \end{minipage}
  \vspace{5pt}
}
\makeatother

\begin{document}

% \title[R-Eval: Knowledge-oriented Language Agents Assessment]{D-Eval: Benchmarking Domain Knowledge of Retrieval Augmented Large Language Models}
\title[R-Eval: A Unified Toolkit for Evaluating Retrieval Augmented Large Language Models]{R-Eval: A Unified Toolkit for Evaluating Domain Knowledge of Retrieval Augmented Large Language Models}

\author{Shangqing Tu}
\authornote{Both authors contributed equally to this research.}
\affiliation{%
  \institution{DCST, Tsinghua Univerisity}
  \city{Beijing 100084}
  \country{China}
}
\email{tsq22@mails.tsinghua.edu.cn}

\author{Yuanchun Wang}
\authornotemark[1]
\affiliation{%
  \institution{SoI, Renmin University of China}
  \city{Beijing 100084}
  \country{China}
}
\email{wangyuanchun@ruc.edu.cn}

\author{Jifan Yu}
\affiliation{%
  \institution{DCST, Tsinghua Univerisity}
  \city{Beijing 100084}
  \country{China}
}
\email{yujf21@mails.tsinghua.edu.cn}

\author{Yuyang Xie}
\affiliation{%
  \institution{DCST, Tsinghua Univerisity}
  \city{Beijing 100084}
  \country{China}
}
\email{xieyy21@mails.tsinghua.edu.cn}

\author{Yaran Shi}
\affiliation{%
  \institution{SIOE, Beihang Univerisity}
  \city{Beijing 100084}
  \country{China}
}
\email{syr2021@buaa.edu.cn}

\author{Xiaozhi Wang}
\affiliation{%
  \institution{DCST, Tsinghua Univerisity}
  \city{Beijing 100084}
  \country{China}
}
\email{wangxz20@mails.tsinghua.edu.cn}

% \author{Chunyang Li}
% % \authornote{Both authors contributed equally to this research.}
% \affiliation{%
%   \institution{DCST, Tsinghua Univerisity}
%   \city{Beijing 100084}
%   \country{China}
% }
% \email{yaozj20@mails.tsinghua.edu.cn}

% \author{Daniel Zhang-Li}
% \affiliation{%
%   \institution{BNRist, DCST, Tsinghua Univerisity}
%   \city{Beijing 100084}
%   \country{China}
% }
% \email{jietang@tsinghua.edu.cn}

% \author{Zheyuan Zhang}
% % \authornote{Both authors contributed equally to this research.}
% \affiliation{%
%   \institution{IoE, Tsinghua Univerisity}
%   \city{Beijing 100084}
%   \country{China}
% }
% \email{marylee@mail.tsinghua.edu.cn}

\author{Jing Zhang}
\authornote{Corresponding authors.}
\affiliation{%
  \institution{SoI, Renmin University of China}
  \city{Beijing 100084}
  \country{China}
}
\email{zhang-jing@ruc.edu.cn}

\author{Lei Hou}
\authornotemark[2]
\affiliation{%
  \institution{BNRist, DCST, Tsinghua Univerisity}
  \city{Beijing 100084}
  \country{China}
}
\email{houlei@tsinghua.edu.cn}

\author{Juanzi Li}
\affiliation{%
  \institution{BNRist, DCST, Tsinghua Univerisity}
  \city{Beijing 100084}
  \country{China}
}
\email{lijuanzi@tsinghua.edu.cn}

\renewcommand{\shortauthors}{Tu and Wang, et al.}

\begin{abstract}

Large language models have achieved remarkable success on general NLP tasks, but they may fall short for domain-specific problems. Recently, various Retrieval-Augmented Large Language Models (RALLMs) are proposed to address this shortcoming. However, existing evaluation tools only provide a few baselines and evaluate them on various domains without mining the depth of domain knowledge. In this paper, we address the challenges of evaluating RALLMs by introducing the R-Eval toolkit, a Python toolkit designed to streamline the evaluation of different RAG workflows in conjunction with LLMs.  Our toolkit, which supports popular built-in RAG workflows and allows for the incorporation of customized testing data on the specific domain, is designed to be user-friendly, modular, and extensible. We conduct an evaluation of 21 RALLMs across three task levels and two representative domains, revealing significant variations in the effectiveness of RALLMs across different tasks and domains. Our analysis emphasizes the importance of considering both task and domain requirements when choosing a RAG workflow and LLM combination. We are committed to continuously maintaining our platform at \url{https://github.com/THU-KEG/R-Eval} to facilitate  both the industry and the researchers.

\end{abstract}

\begin{CCSXML}
<ccs2012>
   <concept>
       <concept_id>10010147.10010178.10010179.10010181</concept_id>
       <concept_desc>Computing methodologies~Discourse, dialogue and pragmatics</concept_desc>
       <concept_significance>500</concept_significance>
       </concept>
   <concept>
       <concept_id>10010147.10010178.10010179.10010182</concept_id>
       <concept_desc>Computing methodologies~Natural language generation</concept_desc>
       <concept_significance>500</concept_significance>
       </concept>
 </ccs2012>
\end{CCSXML}

% \ccsdesc[500]{Applied computing~Computer-assisted instruction}
\ccsdesc[500]{Computing methodologies~Discourse, dialogue and pragmatics}
\ccsdesc[500]{Computing methodologies~Natural language generation}

\keywords{Evaluation, Domain Knowledge, Large Language Model}

\maketitle

% These models exhibit an unparalleled capability to understand and synthesize segments of both programming and natural languages, enabling them to follow general instructions and solve complex problems~\citep{chia2023instructeval,frieder2023mathematical}. , in the training data of these general-purpose LLMs

\section{Introduction}

The burgeoning advancements in large language models (LLMs), such as GPT-4~\citep{openai2023gpt4} and Llama~\citep{touvron2023llama}, have sparked widespread interest across industry, academia, and the public sphere~\citep{bubeck2023sparks}. However, while LLMs excel in general tasks~\citep{chia2023instructeval,frieder2023mathematical}, their performance can falter when confronted with domain-specific tasks~\citep{kocon2023chatgpt,yang2023harnessing}. This shortfall is primarily due to the potential absence of domain knowledge~\cite{anand1995role}, defined as pre-existing information and expertise within a specific field. Consequently, the technique of retrieval augmented generation (RAG)~\cite{lewis2020retrieval} has gained traction, particularly in adapting LLMs for domain-specific applications such as AI healthcare assistants, as it offers a solution to mitigate the propensity of LLMs to generate hallucinated responses~\citep{HaluEval,tonmoy2024comprehensive,wu2023ragtruth}.

\begin{figure}[t]
    \centering
    \includegraphics[width=\linewidth]{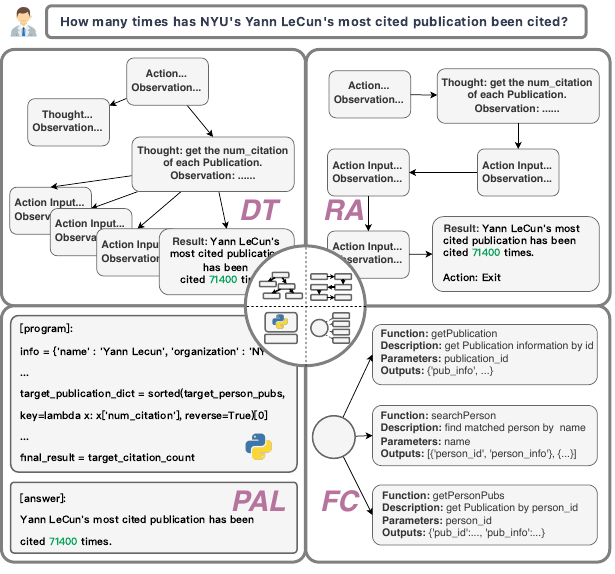}
    \caption{Responses from $4$ RAG workflows, including DFSDT(DT)~\cite{toolllm}, ReAct(RA)~\cite{yao2022react}, PAL~\citep{gao2023pal} and GPT Function Calling (FC) for a domain-specific question.}
    \label{fig:intro}
\end{figure}

As LLMs evolve~\citep{yu2023kola}, becoming more powerful and resource-intensive than small pre-trained language models, a plethora of new RAG workflows have been introduced specifically for these models. As depicted in Figure~\ref{fig:intro}, these innovative approaches typically start by retrieving relevant resources based on user input. Subsequently, they either synthesize outputs from independently executed tools, as exemplified by program-aided language model workflow (PAL)~\citep{gao2023pal} and OpenAI's Function Calling\footnote{\url{https://platform.openai.com/docs/guides/function-calling}\label{fc}} (Figure~\ref{fig:intro} bottom), or adopt a sequential execution and prompt-based reasoning process, as demonstrated by DFSDT~\cite{toolllm} and ReAct~\citep{yao2022react} (Figure~\ref{fig:intro} top). These advancements in RAG workflows for LLMs are opening new avenues for exploration and evaluation.

Despite the progress in the field, there is still a notable scarcity of software tools that offer a simple, all-in-one evaluation system for different retrieval augmented large language models (RALLMs). However, prior evaluation works lacks consideration of two key factors: (1) Insufficient exploration of \textbf{combinations between LLMs and RAG workflows}. Recent RAG evaluation frameworks  such as RAGAS~\citep{es2023ragas} and ARES~\citep{saad2023ares}, offer only a limited number of baselines and do not fully explore the myriad possible combinations of RAG workflows and LLMs~\citep{liu2023bolaa}. (2) Lack comprehensive \textbf{mining of the domain knowledge}. Furthermore, recent benchmarks like  ToolBench~\citep{toolllm} and API-Bank~\citep{li-etal-2023-api} primarily focus on the general capabilities of RALLMs across various domains, with each domain typically containing only about 10 test cases on average, resulting in a shortage of testing data. A unified toolkit could facilitate fair comparisons and promote wider adoption of various RALLM systems for domain-specific applications.

To address this gap, we introduce R-Eval (RALLM Evaluation Toolkit), a Python toolkit designed to streamline the evaluation of different RAG workflows in conjunction with LLMs. Our toolkit offers users the flexibility to explore four popular built-in RAG workflows: ReAct~\citep{yao2022react}, PAL~\citep{gao2023pal}, DFSDT~\citep{toolllm}, and function calling\footref{fc}. In addition, it provides the capability to incorporate customized testing data in specific domains through template-based question generation. An included analysis module can automatically perform performance, error, and deployment analyses. Distinct from other LLM evaluation toolkits, R-Eval is one of the first to prioritize the evaluation of domain knowledge in RALLMs. It is designed to be user-friendly, modular, and extensible, offering a robust and comprehensive evaluation framework for the community.

Based on the R-Eval toolkit, we conduct an evaluation of $21$ RALLMs  for three task levels on two representative domains: Wiki-pedia~\citep{yang2015wikiqa} and Aminer~\citep{aminer}, leading to some interesting findings:

We discover that the effectiveness of RALLMs can vary significantly across different tasks and domains. In particular, the combination of the ReAct workflow with the GPT-4-1106 model exhibits exceptional performance across all tasks and domains, making it a robust choice for a variety of applications. However, the optimal RAG workflow and LLM combination may vary depending on the specific task or domain. For instance, other combinations such as PAL with GPT-4-1106 or ReAct with Llama2-7B-chat~\citep{touvron2023llama} also demonstrate strong performance in certain areas. Our Matching Analysis further reveals that while GPT-4-1106 is the best matching LLM for the ReAct workflow across all tasks and domains, other workflows like PAL may yield comparable results with different LLMs, such as GPT-3.5-turbo-1106.

In our Error Analysis, we find that the PAL workflow produces the highest proportion of tool-using errors, indicating that it often fails to successfully retrieve domain knowledge, which could explain why GPT-4 performs poorly with PAL. On the other hand, workflows like ReAct, DFSDT, and FC generate fewer tool-using errors, resulting in more successful knowledge retrieval and better overall performance. In terms of practical deployment, we find that GPT models, which are called through APIs, ``outperform'' open-source LLMs in terms of both efficiency and effectiveness. Among the open-source LLMs, Tulu-7B~\citep{wang2023far} strikes a good balance between efficiency and effectiveness, showcasing its potential for practical applications. Our findings underscore the importance of considering both task and domain requirements when choosing a RAG workflow and LLM combination. These insights offer valuable guidance for developers and researchers in the selection of suitable RALLMs for specific tasks or domains.

\textbf{Impact and Beneficiaries.} We've crafted the R-Eval toolkit for thorough RALLMs evaluations and performance analysis. This toolkit aids: (1) \textit{Researchers} in evaluating and contrasting RALLMs across tasks and domains, thereby guiding future research. (2) \textit{Industry Professionals}, particularly in AI, by offering a resource for assessing RALLMs' real-world applicability, aiding in informed model selection. (3) \textit{Developers}, by providing a flexible platform to test, refine, and deploy their RALLMs, and understand trade-offs between efficiency and effectiveness.

 % \textbf{Impact and Beneficial Groups.} In summary, we systematically design the R-Eval toolkit to conduct comprehensive evaluations for RALLMs and analyse its performance on specific domains. We hope our work can benefit the following groups: ((1) \textit{Researchers}: Our toolkit provides a robust platform for researchers to evaluate and compare various RALLMs across different tasks and domains. By revealing the strengths and weaknesses of each model, it can help researchers identify areas of improvement and guide future research directions.  (2) \textit{Industry Professionals}: For industry professionals, particularly those in the field of AI, our toolkit can serve as a valuable resource for assessing the applicability and effectiveness of different RALLMs in real-world scenarios. The insights gained from the toolkit can help professionals make informed decisions about which models to adopt or invest in, depending on their specific use-cases and domains. (3) \textit{Developers}: The R-Eval toolkit offers a user-friendly, modular, and extensible platform for developers to test and refine their RALLMs. It can help developers understand the performance characteristics of their models, identify and rectify errors, and optimize their models for specific tasks or domains. Additionally, the toolkit can assist developers in the practical deployment of RALLMs, providing insights into the trade-offs between efficiency and effectiveness.

% \section{Preliminaries}

\begin{figure*}[htbp]
    \centering
    \includegraphics[width=\linewidth]{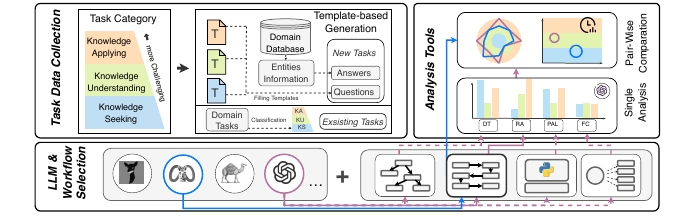}
    \caption{%The framework of our ext-then-abs model on multi-document summarization. 
    The framework of our evaluating and analysing process. We first choose an environment and collect testing data from both existing benchmarks and template-based QA pairs. Then we select the RAG workflows and LLMs to form a RALLM for running the evaluation. After that, we perform a comprehensive analysis on results to get insights.
    }
    \label{fig:pipeline}
\end{figure*}

\section{Background}
\label{sec:background}

\textbf{Retrieval Augmented LLMs}.  Since LLMs have successfully improved the performance of various general domain tasks~\cite{bubeck2023sparks,openai2023gpt4}, researchers have been exploring the methods that adapt LLMs to domain-specific tasks. The most common solution is to build a retrieval augmented LLM (RALLM) that utilize the domain knowledge via retrieval~\cite{liu2023reta}. Previous retrieval workflows for LLM can be broadly categorized into two categories: (1) Planned Retrieval: The retriever plans what knowledge to fetch according to the question and pass them to LLMs for the answer generation~\cite{li2022decoupled,lazaridou2022internet,shi2023replug,liu2023lost}. (2) Interactive Retrieval: As the retriever may occasionally fail to yield accurate or comprehensive results~\cite{wang2023survey}, an interactive retrieval mechanism that grants LLMs the capability to refine the retrieval process can effectively tackle these challenges~\cite{yao2022react,trivedi-etal-2023-interleaving,jiang-etal-2023-active,varshney2023stitch,shinn2023reflexion}.

\textbf{Domain-specific Evaluation}.  The powerful generation ability of large language models (LLMs) has had a profound impact in various fields, such as law~\cite{fei2023lawbench}, finance~\cite{xie2023pixiu}, medicine~\cite{li2023huatuo,jin2023genegpt} and education~\cite{fan2023grammargpt,liu2023argugpt}. Therefore, many efforts have been devoted to evaluating the capabilities of large language models for a certain domain with various new benchmarks, such as LegalBench~\cite{guha2023legalbench}, CMB~\cite{wang2023cmb} and EcomInstruct~\cite{li2023ecomgpt}. However, previous domain-specific evaluations mainly focus on providing tailored task data, ignoring the importance of building an interactive environment for RAG settings~\cite{lewis2020retrieval}, which are widely adopted in real-world applications. This limitation inspires our work to build easy-to-adapt environments and free-to-combine retrieval workflows for RALLM evaluations.

\textbf{Problem Definition} We define the problem as follows: Given a domain-specific task $T$ that necessitates specialized knowledge from a particular field, an environment $K$ for querying domain knowledge, and a RALLM $R$ with a workflow to fetch relevant information based on the input and generate a response via LLM,  our objective is to assess the performance of $R$ on $T$ using $K$. The main challenge lies in the fact that different combinations of large language models and retrieval workflows can lead to different results, and the optimal combination may vary depending on the specific task or domain. Furthermore, the evaluation should not only consider the final output but also the process of retrieving information and generating the response, which involves analyzing the strengths and weaknesses of different RALLMs. %Our introduced toolkit, R-Eval, aims to facilitate the systematic evaluation of various RALLMs on diverse domain-specific tasks.

\section{System}
\label{sec:sys}
To investigate how well LLMs can master domain knowledge, we propose R-Eval, a unified evaluation toolkit that can be adapted to any specific domain. As shown in Figure ~\ref{fig:pipeline}, R-Eval can generate test cases, load various RALLMs and analyse their performances.

\subsection{Environment Setting}
\label{sec:env_setting}
In the context of a Retrieval Augmented Large Language Model system, the LLM interacts with the domain's API in a tool-using manner, retrieving the necessary knowledge to accomplish related tasks. In order for LLMs to retrieve domain knowledge, we design an environment with several query APIs for each domain. We begin with two representative domains as use cases: (1) \textit{Wikipedia}\footnote{\url{https://www.wikipedia.org}} is a high-quality knowledge source with over $6.6$ million English articles, which has provided supporting facts for building various open-domain QA tasks~\cite{lewis2020retrieval}. We use a simple Wikipedia web API with three types of functions to enable interactive information retrieval, including Search, Lookup and Finish, which is the same as ReAct's environment~\cite{yao2022react}. (2) \textit{Aminer}\footnote{\url{https://www.aminer.cn}}~\cite{aminer} is an academic information service system containing over $69$ million scholar profiles and $290$ million publications. We define 7 query APIs for data retrieving about scholars and publications, including searchPerson, searchPublication and getCoauthors, etc (More are in Appendix ~\ref{sec:implementation}).

\begin{table*}[t!]
\centering
\caption{The domain specific tasks in R-Eval. Test Set and Pool correspond to the testing instances used in each season and the overall available instances. \emph{Existing} tasks means their test sets are taken from the original dataset. \emph{Refreshing} tasks' instances are newly developed by our template-based question generation process.  }
\label{tab:tasks}
\resizebox{0.85\linewidth}{!}{
\begin{tabular}{@{}c|c|c|l|cccc|c@{}}
\toprule
 \textbf{Domain}  &  \textbf{Level}  &   \textbf{ID}             & \textbf{Dataset}              & \textbf{Metrics}  & \textbf{Context Type} & \textbf{Test Set} & \textbf{Pool} & \textbf{Source}            \\ \midrule
 \multirow{9}{*}{Wiki} &  \multirow{2}{*}{KS} & 1-1           & High-Freq.      &    \textbf{F1}                            &    Triple                &       100               &        $20.6$M            & Refreshing \\
    &    &         1-2                          & Low-Freq.       &    \textbf{F1}                            &       Triple               &       100                &       $20.6$M             &                       Refreshing     \\  \cmidrule{2-9}
 & \multirow{3}{*}{KU}         &          2-1                        & COPEN-CSJ                     &     \textbf{F1}        &       Entity, Concept              &     100                  &  $3.9$k                  &   Existing                         \\
     &    &        2-2                          & COPEN-CPJ                     &      \textbf{F1}                        &       Concept               &       100                 &     $4.7$k               &                      Existing      \\
     &     &          2-3                       & COPEN-CiC                     &         \textbf{F1}             &         Concept             &           100             &        $2.3$k            &                       Existing     \\  \cmidrule{2-9}
 &   \multirow{4}{*}{KA} & 3-1              & HotpotQA                      &        \textbf{F1}               &      Document(s)             &         100               &      $7.4$k              & Existing \\
      &    &           3-2                     & 2WikiMulti.                 &       \textbf{F1}                    &          Document(s)            &       100                 &      $12.6$k              &                     Existing       \\
    &     &         3-3                        & MuSiQue                       &       \textbf{F1}                       &        Document(s)              &      100                  & $2.5$k                   &                     Existing       \\ 
   &      &        3-4                         & KQA Pro                       &        \textbf{F1}                       &          KG            &       100                 &     $1.2$k               &               Refreshing             \\ \midrule
    \multirow{3}{*}{Aminer}   &    KS     &          1-3                        & Soay-Easy  &     \textbf{F1}                  &        Triple            &           100             &         $12.7$k           & Refreshing                     \\ 
    &    KU &            2-4                       & Profiling &              \textbf{F1}                      &    Document, Entity             &             100           &          $1.8$k          & Existing                    \\
    &  KA    &      3-5                          & Soay-Hard      &         \textbf{F1}                       &       KG            &       100                 &      $12.7$k              & Refreshing                   \\  \bottomrule
\end{tabular}}
\end{table*}

\subsection{Task Data Collection}
\label{sec:data_collection}

Knowledge, which means information including facts, events, and skills, has been utilized as an indicator for AI's intelligence level~\cite{newell1982knowledge}.

\paragraph{Task Category.} Considering the cognitive ability modeling for knowledge~\cite{yu2023kola}, we select three widely accepted processes in Bloom's taxonomy~\citep{krathwohl2002revision} for organizing the tasks in R-Eval benchmark:

$\bullet$ \textbf{Knowledge Seeking (KS)} is designed to assess the model's capacity to accurately recall established facts from the  environment, as demonstrated by the prior Open-domain QA task~\citep{chen-etal-2017-reading}.

$\bullet$ \textbf{Knowledge Understanding (KU)} aims to evaluate the model's proficiency in comprehending the inherent knowledge embedded in texts, as manifested by the tasks of information extraction~\citep{yao-etal-2019-docred,peng2022copen}.

$\bullet$ \textbf{Knowledge Application (KA)} measures the ability of models to utilize retrieved knowledge in performing reasoning and problem-solving tasks, which is assessed through various knowledge reasoning tasks like multi-hop reasoning~\citep{yang2018hotpotqa,trivedi2022musique,cao-etal-2022-kqa}.

\paragraph{Test Datasets.}  The R-Eval benchmark, as depicted in Table \ref{tab:tasks}, encompasses 12 tasks designed to assess the three levels of cognitive ability for 2 domains. These tasks are constructed from both existing datasets and  refreshing template-based question generation. 

\textbf{Knowledge Seeking Tasks} evaluate the model's ability to recall facts. The tasks on wiki domain are constructed from triplets in Wikidata5M, transformed into sentences with relation-specific templates. Two test sets are created, one for high-frequency knowledge (1-1) and another for low-frequency knowledge (1-2). A knowledge seeking test of aminer domain (1-3) is also included, which annotates knowledge triplets from the aminer knowledge base.

\textbf{Knowledge Understanding Tasks} assess the model's understanding of various types of knowledge from texts. They include Concept Probing (2-1/2-2/2-3) from the COPEN dataset for Wikipedia, which are conceptual similarity judgment (CSJ), conceptual property judgment (CPJ) and conceptualization in contexts (CiC) respectively.  A dataset on aminer (2-8) is also included, which evaluates the ability of extracting structured profiles from plain text.

\textbf{Knowledge Application Tasks}  measure the model's ability to apply retrieved knowledge in multi-hop reasoning tasks. They include tasks from the HotpotQA (3-1), 2WikiMultihopQA (3-2), MuSiQue (3-3), and KQA Pro (3-4) datasets. From 3-1 to 3-4, the reasoning difficulty on wiki domain knowledge is increasing. Another KA test for aminer domain (3-6) is also included, producing questions based on the domain knowledge.

% As shown in Table \ref{tab:tasks}, various existing datasets are utilized to evaluate the three levels of cognitive abilities in the R-Eval benchmark.

% For the \textbf{Knowledge Seeking (KS)} level, the Soay-Easy dataset from the Aminer domain is used. This dataset is designed to assess the model's ability to accurately recall established facts from the environment.

% In the \textbf{Knowledge Understanding (KU)} level, three datasets from the Wiki domain (COPEN-CSJ, COPEN-CPJ, and COPEN-CiC) and one from the Aminer domain (Profiling) are used. These datasets focus on the model's proficiency in comprehending inherent knowledge embedded in texts, particularly in the context of entities and concepts.

% For the \textbf{Knowledge Application (KA)} level, three datasets from the Wiki domain (HotpotQA, 2WikiMulti, and MuSiQue) and one from the Aminer domain (Soay-Hard) are employed. These datasets measure the ability of models to utilize retrieved knowledge in performing reasoning and problem-solving tasks, particularly in the context of documents and knowledge graphs.

% 元淳
\paragraph{Template-based Generation.} Inspired by previous works~\citep{cao-etal-2022-kqa, wang2024soay}, we utilize a template-based generation approach to rapidly construct evaluation sets from a given domain database. More specifically, we initially compose a series of template questions with placeholders based on the content we need to evaluate, then fill these templates with meaningful information by sampling entries from the domain database. Additionally, the answers to these questions can be also derived directly from the entity information. Notably, after manually crafting a few template questions, LLMs can be employed to quickly complete subsequent template construction.

Taking the domain of academic intelligence as an example, we could design a template question such as \textit{"What are the research interests of XXX at X institution?"}. From the AMiner database, we can extract a plethora of scholar entity information like names, affiliations, interests, and email addresses, for example, \textit{\{`name' : `Yann Lecun', `organization' : 'New York University', `Interest' : `AI, Machine Learning, Computer Vision, Robotics, Image Compression', `email' : `yl22@nyu.edu', …\}}. Consequently, this template question can be filled as, \textit{"What are the research interests of Yann Lecun at New York University?"} and the answer can be found within the entity information.
Employing this method, we've designed an extensive range of QA test sets for distinct task difficulty levels. The \textit{aminer KS} and \textit{aminer KA} tasks shown in table~\ref{tab:aminer_res} are generated in this way with varying difficulty levels.

\begin{table*}[t]
  \centering
    \caption{Comparison with existing works' evaluations of RALLMs. \# is the short for `number'. The column \textit{Levels of Tasks} refers to the task categories divided in the evaluation. \textit{Analysis Toolkit} means whether this work provides an analysis toolkit.}
  % \scriptsize
  \begin{tabular}{cccccccc}
  \toprule
  \multicolumn{1}{c}{Research for RALLM}  & \begin{tabular}[c]{@{}c@{}}  Number of\\  RAG workflows\end{tabular} & \begin{tabular}[c]{@{}c@{}}  Number\\  of LLMs  \end{tabular} & \begin{tabular}[c]{@{}c@{}} \# of evaluated\\   
feasible RALLMs  \end{tabular} & \begin{tabular}[c]{@{}c@{}}  Levels\\  of Tasks
  \end{tabular} & \begin{tabular}[c]{@{}c@{}}  Number\\  of Tasks  \end{tabular} & \begin{tabular}[c]{@{}c@{}}  Avg. test data\\  for each domain  \end{tabular}  & \begin{tabular}[c]{@{}c@{}} Analysis\\  Toolkit
  \end{tabular}  \\ 
  \midrule
     ToolQA~\cite{zhuang2023toolqa}   & 2 & 2   & 3   &   (easy, hard) & 6 & 255  &  $\times$  \\
  ToolBench~\cite{toolllm}    &  3 & 5  & 6   & (I1, I2, I3)  & 6  & 258.3  & $\times$  \\
  API-Bank~\cite{li-etal-2023-api}      &  1 & 6   & 6  & None  & 4 & 6.1  & $\times$   \\
   % BOLAA~\cite{liu2023bolaa}      &  6 & 15  & 90    & 2  & Task complexity & 300 & $\times$  \\

% RALLE~\cite{hoshi-etal-2023-ralle}     & 2 & 3    & 4     & 1  &  51,464 & None & $\times$   \\
FLARE~\cite{jiang-etal-2023-active}     & 4 & 1    & 4     & None  &  4 & 432.3 & $\times$   \\

  IRCoT~\cite{trivedi-etal-2023-interleaving}   & 3 & 2   & 6   & None  &   4 & 500  & $\times$   \\
  DECOMP~\cite{khot2022decomposed}   & 3 & 2   & 6   & None &   4 & 100  & $\times$   \\
ReAct~\cite{yao2022react}   & \textbf{4}  & 3   & 12 & None   &  2 & 500  & $\times$  \\
  \midrule
  \textbf{R-Eval (ours)}  & \textbf{4} & \textbf{8} & \textbf{21} & (KS, KU, KA)  & \textbf{12}  & \textbf{600}  & $\checkmark$ \\ 
  \bottomrule
  \end{tabular}

   \label{tab:comp}
  \end{table*}

\subsection{Workflow and LLM Selection}

As shown in Table~\ref{tab:comp}, a RALLM system usually composes of a workflow for retrieving domain knowledge and an LLM for reasoning~\cite{liu2023reta}. R-Eval presents a feasible  protocol that combines different workflows and LLMs for comprehensive evaluation.

\paragraph{Workflow Selection.} Following BOLAA~\cite{liu2023bolaa}, we select $4$ typical baseline workflows for retrieving knowledge from our environment: (1) \textit{ReAct}~\cite{yao2022react} is a prompt-based paradigm designed to synergize reasoning and acting in language models for task solving, creating action spaces for LLMs to retrieve external knowledge. (2) \textit{PAL}~\cite{gao2023pal} is a program-aided workflow that allows LLMs to generating python programs for executing API queries. (3) \textit{DFSDT}~\cite{qin2023toolllm} is a general decision-making strategy for API usage, enhancing the reasoning capabilities of large language models (LLMs) by considering multiple reasoning traces in the environment. (4) \textit{FC}\footref{fc} is short for Function Calling, which is a close-source tool using workflow by OpenAI.

\paragraph{LLM Selection.} 
The evaluated LLMs consist of two categories: (1) \emph{Open-source Model}, including Llama2-chat (7B)~\citep{touvron2023llama}, Tulu (7B)~\citep{wang2023far}, Vicuna (13B)~\citep{zheng2023judging}, Llama2 (13B)~\citep{touvron2023llama}, CodeLlama-instruct (13B)~\citep{roziere2023code}, ToolLlama-2 (7B)~\citep{toolllm}. (2)  \emph{Commercial Model}: GPT3.5-turbo\footnote{\url{https://platform.openai.com/overview}\label{openai}} and GPT-4~\citep{openai2023gpt4}.
As the workflows may require LLMs having the instruction following abilities to produce specific output format like code for PAL, not all LLMs can fit the workflows. Therefore, we tried to combine LLMs with the workflows and keep those combination that can achieve non-zero performance on our benchmark. Finally, we get $21$ feasible RALLM systems for evaluation, which has more RALLMs than those implemented in previous works~\citep{toolllm,yao2022react}.

% 在特定domain上获取了不同RALLM系统的任务表现之后，我们对结果进行了分析。我们开发了一套自动化的分析工具： (1) Single System Analysis: 有对单个系统在不同类型任务上的表现分析，还有对其在具体任务上的错误类型占比分析。 (2) Pairwise System Analysis: 这部分是对不同系统之间的比较和ensemble分析，我们首先以某个系统的得分为基准，normalize其他系统的表现以比较他们的水平差异。其次，我们尝试了对任意数量的系统做best-of-n的ensemble，计算他们在blender的情况下能够获得多高的表现。 (3) Depolyment Analysis: 展现了不同系统在运行时的速度与效果之间的trade-off，为选型提供了参考。

\subsection{Analysis Tools}
\label{analysis_tools}
We developed three automated analysis tools to evaluate the performance of various RALLMs in domain-specific tasks. (1) \textbf{Matching Analysis} facilitates comparison between systems on the same RAG workflow by visualizing the evaluation results. It normalizes systems' performance via a radar map, which can reflect the difference directly. (2) \textbf{Error Analysis} provides a detailed view of each system's weakness across different tasks and identifies common error types. This helps to pinpoint each system's areas for improvement. (3) \textbf{Deployment Analysis} assesses the trade-off between runtime speed and performance for each system, aiding decision-making in system deployment. These tools offer a comprehensive, multifaceted analysis of RALLM system performance, providing valuable insights for their application in domain-specific tasks.

\paragraph{Error and Response Types.} To provide a more comprehensive error analysis for RALLMs, we have conducted a fine-grained taxonomy of system response types, which include 
all error and correct cases.

Under the R-Eval framework, all RAG systems, besides preserving the final outcome (response), also document the information retrieved (scratchpad), and whether the inference process was interrupted due to erroneous usage or faults with the retrieval tools. Based on these records, we categorized the system responses into six types in a fine-grained manner. When the response aligns with the standard answer, if the scratchpad contributes to the final response, we define this situation as \textbf{E}xact \textbf{M}atch (EM). If the information in the scratchpad doesn't relate to the response, i.e., the system can answer consistently with the standard answer relying on the knowledge stored within the model, it's considered an \textbf{A}nswer \textbf{M}atch (AM). As to the situation that the response does not match the standard answer, if the scratchpad provides useful information, there are no problems with the retrieval process but something is wrong in the model's generation process grounded on the retrieved content, we define this as \textbf{G}rounded-generation \textbf{E}rror (GE). If the scratchpad's information is irrelevant, the faulty retrieval process leads to undesirable responses. We then have to ascertain whether the retrieval process was interrupted. There is an issue with the logic of the retrieval if the process is halted. We denote this as a \textbf{R}easoning \textbf{E}rror (RE). If the retrieval process was interrupted, depending on whether the interruption was due to faults in the retrieval tool itself or improper use of the tool, it is classified as a \textbf{M}odel \textbf{E}rror (ME) or \textbf{T}ool-using \textbf{E}rror (TE), respectively. Specially,  R-Eval uses the F1-Score between system responses and standard answers to determine whether the response matches the standard answer, and the usefulness of the scratchpad is assessed by checking if the response is contained within the scratchpad.

This kind of fine-grained classification enables us to quickly assess the capabilities of an RAG system and diagnose the cause of errors. For instance, a higher proportion of EMs signifies better RAG capabilities. A high proportion of AMs implies the system model has strong internal capabilities but weak retrieval capabilities. GE is in contrast to AM: the higher the proportion of GEs, the stronger the retrieval capabilities but the weaker the model's capabilities. A high proportion of REs might indicate that the current system's reasoning capabilities are insufficient to support the current workflow. The presence of MEs and TEs reveals issues with the model itself or the retrieval tool interface, necessitating separate inspections.

\begin{table*}[ht]
\caption{Performance of different systems for each task on Aminer domain and the average performance of overall tasks. }
\centering
  % \small
% \resizebox{\linewidth}{!}{
\begin{tabular}{l|l|rr|rr|rr|rrrrrr}
\toprule
\multirow{2}{*}{\textbf{Workflow}} & \multirow{2}{*}{\textbf{LLM}} & \multicolumn{2}{c}{\textbf{\cellcolor{'shallow1'} aminer KS}}                                                                                                                                                                                                            & \multicolumn{2}{|c}{\textbf{\cellcolor{'shallow2'} aminer KU}}     
                                                                 & \multicolumn{2}{|c}{\textbf{\cellcolor{'shallow3'} aminer KA}}                                                                & \multicolumn{6}{|c}{\textbf{\cellcolor{'shallow4'} Overall Average (Level 1, 2, 3)}}                                 \\ 
                             &   & \multicolumn{1}{c}{\cellcolor{'deep1'} \textbf{1-3}} & \multicolumn{1}{c}{\cellcolor{'deep1'} \textbf{Rank}} & \multicolumn{1}{|c}{\cellcolor{'deep2'} \textbf{2-4}}  & \multicolumn{1}{c}{\cellcolor{'deep2'} \textbf{Rank}} & \multicolumn{1}{|c}{\cellcolor{'deep3'} \textbf{3-5}}  & \multicolumn{1}{c}{\cellcolor{'deep3'} \textbf{Rank}} & \multicolumn{1}{|c}{\cellcolor{'deep4'} \textbf{wiki}} & \multicolumn{1}{c}{\cellcolor{'deep4'} \textbf{Rank}} & \multicolumn{1}{c}{\cellcolor{'deep4'} \textbf{aminer}} & \multicolumn{1}{c}{\cellcolor{'deep4'} \textbf{Rank}} & \multicolumn{1}{c}{\cellcolor{'deep4'} \textbf{all}} & \multicolumn{1}{c}{\cellcolor{'deep4'} \textbf{Rank}} \\ \midrule
\cellcolor{'wit'} ReAct & \cellcolor{'wit'} gpt-4-1106 & \cellcolor{'shallow1'} 89.7 & \cellcolor{'shallow1'}  1st & \cellcolor{'shallow2'} 46.7 & \cellcolor{'shallow2'}  3rd & \cellcolor{'shallow3'} 57.7 & \cellcolor{'shallow3'}  1st & \cellcolor{'shallow4'}  38.8 & \cellcolor{'shallow4'}  1st & \cellcolor{'shallow4'}  64.7 & \cellcolor{'shallow4'}  1st & \cellcolor{'shallow4'}  45.3 & \cellcolor{'shallow4'}  1st  \\ 
\cellcolor{'gry'} PAL & \cellcolor{'gry'} gpt-3.5-turbo & \cellcolor{'deep1'} 80.1 & \cellcolor{'deep1'}  3rd & \cellcolor{'deep2'} 50.7 & \cellcolor{'deep2'}  2nd & \cellcolor{'deep3'} 54.9 & \cellcolor{'deep3'}  2nd & \cellcolor{'deep4'}  19.9 & \cellcolor{'deep4'}  6th & \cellcolor{'deep4'}  61.9 & \cellcolor{'deep4'}  2nd & \cellcolor{'deep4'}  30.4 & \cellcolor{'deep4'}  2nd  \\ 
\cellcolor{'wit'} PAL & \cellcolor{'wit'} gpt-4-1106 & \cellcolor{'shallow1'} 59.3 & \cellcolor{'shallow1'}  4th & \cellcolor{'shallow2'} 56.8 & \cellcolor{'shallow2'}  1st & \cellcolor{'shallow3'} 52.7 & \cellcolor{'shallow3'}  3rd & \cellcolor{'shallow4'}  20.3 & \cellcolor{'shallow4'}  5th & \cellcolor{'shallow4'}  56.2 & \cellcolor{'shallow4'}  3rd & \cellcolor{'shallow4'}  29.2 & \cellcolor{'shallow4'}  3rd  \\ 
\cellcolor{'gry'} ReAct & \cellcolor{'gry'} llama2-7b-chat & \cellcolor{'deep1'} 45.2 & \cellcolor{'deep1'}  5th & \cellcolor{'deep2'} 36.5 & \cellcolor{'deep2'}  6th & \cellcolor{'deep3'} 21.5 & \cellcolor{'deep3'}  6th & \cellcolor{'deep4'}  23.8 & \cellcolor{'deep4'}  3rd & \cellcolor{'deep4'}  34.4 & \cellcolor{'deep4'}  5th & \cellcolor{'deep4'}  26.4 & \cellcolor{'deep4'}  4th  \\ 
\cellcolor{'wit'} PAL & \cellcolor{'wit'} llama2-13b & \cellcolor{'shallow1'} 25.3 & \cellcolor{'shallow1'}  6th & \cellcolor{'shallow2'} 36.4 & \cellcolor{'shallow2'}  7th & \cellcolor{'shallow3'} 20.3 & \cellcolor{'shallow3'}  7th & \cellcolor{'shallow4'}  25.2 & \cellcolor{'shallow4'}  2nd & \cellcolor{'shallow4'}  27.3 & \cellcolor{'shallow4'}  6th & \cellcolor{'shallow4'}  25.7 & \cellcolor{'shallow4'}  5th  \\ 
\cellcolor{'gry'} ReAct & \cellcolor{'gry'} gpt-3.5-turbo & \cellcolor{'deep1'} 84.6 & \cellcolor{'deep1'}  2nd & \cellcolor{'deep2'} 4.0 & \cellcolor{'deep2'}  14th & \cellcolor{'deep3'} 33.0 & \cellcolor{'deep3'}  4th & \cellcolor{'deep4'}  19.6 & \cellcolor{'deep4'}  7th & \cellcolor{'deep4'}  40.6 & \cellcolor{'deep4'}  4th & \cellcolor{'deep4'}  24.9 & \cellcolor{'deep4'}  6th  \\ 
\cellcolor{'wit'} ReAct & \cellcolor{'wit'} vicuna-13b & \cellcolor{'shallow1'} 19.9 & \cellcolor{'shallow1'}  10th & \cellcolor{'shallow2'} 6.0 & \cellcolor{'shallow2'}  13th & \cellcolor{'shallow3'} 7.1 & \cellcolor{'shallow3'}  16th & \cellcolor{'shallow4'}  20.7 & \cellcolor{'shallow4'}  4th & \cellcolor{'shallow4'}  11.0 & \cellcolor{'shallow4'}  17th & \cellcolor{'shallow4'}  18.2 & \cellcolor{'shallow4'}  7th  \\ 
\cellcolor{'gry'} PAL & \cellcolor{'gry'} tulu-7b & \cellcolor{'deep1'} 9.1 & \cellcolor{'deep1'}  15th & \cellcolor{'deep2'} 26.8 & \cellcolor{'deep2'}  9th & \cellcolor{'deep3'} 11.5 & \cellcolor{'deep3'}  12th & \cellcolor{'deep4'}  18.9 & \cellcolor{'deep4'}  8th & \cellcolor{'deep4'}  15.8 & \cellcolor{'deep4'}  9th & \cellcolor{'deep4'}  18.1 & \cellcolor{'deep4'}  8th  \\ 
\cellcolor{'wit'} PAL & \cellcolor{'wit'} vicuna-13b & \cellcolor{'shallow1'} 4.5 & \cellcolor{'shallow1'}  17th & \cellcolor{'shallow2'} 40.9 & \cellcolor{'shallow2'}  4th & \cellcolor{'shallow3'} 2.3 & \cellcolor{'shallow3'}  20th & \cellcolor{'shallow4'}  16.7 & \cellcolor{'shallow4'}  9th & \cellcolor{'shallow4'}  15.9 & \cellcolor{'shallow4'}  8th & \cellcolor{'shallow4'}  16.5 & \cellcolor{'shallow4'}  9th  \\ 
\cellcolor{'gry'} ReAct & \cellcolor{'gry'} llama2-13b & \cellcolor{'deep1'} 16.7 & \cellcolor{'deep1'}  13th & \cellcolor{'deep2'} 0.7 & \cellcolor{'deep2'}  19th & \cellcolor{'deep3'} 23.2 & \cellcolor{'deep3'}  5th & \cellcolor{'deep4'}  15.0 & \cellcolor{'deep4'}  10th & \cellcolor{'deep4'}  13.5 & \cellcolor{'deep4'}  12th & \cellcolor{'deep4'}  14.6 & \cellcolor{'deep4'}  10th  \\ 
\cellcolor{'wit'} PAL & \cellcolor{'wit'} llama2-7b-chat & \cellcolor{'shallow1'} 18.7 & \cellcolor{'shallow1'}  12th & \cellcolor{'shallow2'} 2.8 & \cellcolor{'shallow2'}  15th & \cellcolor{'shallow3'} 16.1 & \cellcolor{'shallow3'}  8th & \cellcolor{'shallow4'}  12.4 & \cellcolor{'shallow4'}  11th & \cellcolor{'shallow4'}  12.5 & \cellcolor{'shallow4'}  14th & \cellcolor{'shallow4'}  12.4 & \cellcolor{'shallow4'}  11th  \\ 
\cellcolor{'gry'} PAL & \cellcolor{'gry'} codellama-13b & \cellcolor{'deep1'} 4.4 & \cellcolor{'deep1'}  18th & \cellcolor{'deep2'} 38.3 & \cellcolor{'deep2'}  5th & \cellcolor{'deep3'} 8.1 & \cellcolor{'deep3'}  14th & \cellcolor{'deep4'}  10.0 & \cellcolor{'deep4'}  14th & \cellcolor{'deep4'}  16.9 & \cellcolor{'deep4'}  7th & \cellcolor{'deep4'}  11.7 & \cellcolor{'deep4'}  12th  \\ 
\cellcolor{'wit'} PAL & \cellcolor{'wit'} toolllama2-7b & \cellcolor{'shallow1'} 1.6 & \cellcolor{'shallow1'}  20th & \cellcolor{'shallow2'} 24.4 & \cellcolor{'shallow2'}  10th & \cellcolor{'shallow3'} 4.6 & \cellcolor{'shallow3'}  18th & \cellcolor{'shallow4'}  12.2 & \cellcolor{'shallow4'}  12th & \cellcolor{'shallow4'}  10.2 & \cellcolor{'shallow4'}  18th & \cellcolor{'shallow4'}  11.7 & \cellcolor{'shallow4'}  13th  \\ 
\cellcolor{'gry'} ReAct & \cellcolor{'gry'} tulu-7b & \cellcolor{'deep1'} 4.0 & \cellcolor{'deep1'}  19th & \cellcolor{'deep2'} 27.8 & \cellcolor{'deep2'}  8th & \cellcolor{'deep3'} 7.9 & \cellcolor{'deep3'}  15th & \cellcolor{'deep4'}  10.3 & \cellcolor{'deep4'}  13th & \cellcolor{'deep4'}  13.2 & \cellcolor{'deep4'}  13th & \cellcolor{'deep4'}  11.0 & \cellcolor{'deep4'}  14th  \\ 
\cellcolor{'wit'} DFSDT & \cellcolor{'wit'} gpt-4-1106 & \cellcolor{'shallow1'} 20.6 & \cellcolor{'shallow1'}  9th & \cellcolor{'shallow2'} 9.6 & \cellcolor{'shallow2'}  12th & \cellcolor{'shallow3'} 11.8 & \cellcolor{'shallow3'}  11th & \cellcolor{'shallow4'}  9.9 & \cellcolor{'shallow4'}  15th & \cellcolor{'shallow4'}  14.0 & \cellcolor{'shallow4'}  11th & \cellcolor{'shallow4'}  10.9 & \cellcolor{'shallow4'}  15th  \\ 
\cellcolor{'gry'} FC & \cellcolor{'gry'} gpt-4-1106 & \cellcolor{'deep1'} 24.7 & \cellcolor{'deep1'}  7th & \cellcolor{'deep2'} 10.9 & \cellcolor{'deep2'}  11th & \cellcolor{'deep3'} 10.2 & \cellcolor{'deep3'}  13th & \cellcolor{'deep4'}  8.2 & \cellcolor{'deep4'}  18th & \cellcolor{'deep4'}  15.3 & \cellcolor{'deep4'}  10th & \cellcolor{'deep4'}  9.9 & \cellcolor{'deep4'}  16th  \\ 
\cellcolor{'wit'} FC & \cellcolor{'wit'} gpt-3.5-turbo & \cellcolor{'shallow1'} 19.0 & \cellcolor{'shallow1'}  11th & \cellcolor{'shallow2'} 1.0 & \cellcolor{'shallow2'}  17th & \cellcolor{'shallow3'} 15.9 & \cellcolor{'shallow3'}  9th & \cellcolor{'shallow4'}  8.8 & \cellcolor{'shallow4'}  16th & \cellcolor{'shallow4'}  12.0 & \cellcolor{'shallow4'}  15th & \cellcolor{'shallow4'}  9.6 & \cellcolor{'shallow4'}  17th  \\ 
\cellcolor{'gry'} ReAct & \cellcolor{'gry'} toolllama2-7b & \cellcolor{'deep1'} 15.0 & \cellcolor{'deep1'}  14th & \cellcolor{'deep2'} 2.2 & \cellcolor{'deep2'}  16th & \cellcolor{'deep3'} 5.7 & \cellcolor{'deep3'}  17th & \cellcolor{'deep4'}  8.3 & \cellcolor{'deep4'}  17th & \cellcolor{'deep4'}  7.6 & \cellcolor{'deep4'}  19th & \cellcolor{'deep4'}  8.1 & \cellcolor{'deep4'}  18th  \\ 
\cellcolor{'wit'} DFSDT & \cellcolor{'wit'} gpt-3.5-turbo & \cellcolor{'shallow1'} 20.7 & \cellcolor{'shallow1'}  8th & \cellcolor{'shallow2'} 0.2 & \cellcolor{'shallow2'}  20th & \cellcolor{'shallow3'} 13.8 & \cellcolor{'shallow3'}  10th & \cellcolor{'shallow4'}  4.8 & \cellcolor{'shallow4'}  20th & \cellcolor{'shallow4'}  11.6 & \cellcolor{'shallow4'}  16th & \cellcolor{'shallow4'}  6.5 & \cellcolor{'shallow4'}  19th  \\ 
\cellcolor{'gry'} ReAct & \cellcolor{'gry'} codellama-13b & \cellcolor{'deep1'} 0.2 & \cellcolor{'deep1'}  21th & \cellcolor{'deep2'} 0.8 & \cellcolor{'deep2'}  18th & \cellcolor{'deep3'} 0.7 & \cellcolor{'deep3'}  21th & \cellcolor{'deep4'}  7.0 & \cellcolor{'deep4'}  19th & \cellcolor{'deep4'}  0.6 & \cellcolor{'deep4'}  21th & \cellcolor{'deep4'}  5.4 & \cellcolor{'deep4'}  20th  \\ 
\cellcolor{'wit'} DFSDT & \cellcolor{'wit'} toolllama2-7b & \cellcolor{'shallow1'} 7.1 & \cellcolor{'shallow1'}  16th & \cellcolor{'shallow2'} 0.0 & \cellcolor{'shallow2'}  21th & \cellcolor{'shallow3'} 2.3 & \cellcolor{'shallow3'}  19th & \cellcolor{'shallow4'}  3.5 & \cellcolor{'shallow4'}  21th & \cellcolor{'shallow4'}  3.1 & \cellcolor{'shallow4'}  20th & \cellcolor{'shallow4'}  3.4 & \cellcolor{'shallow4'}  21th  \\

\bottomrule
\end{tabular}
\label{tab:aminer_res}
\end{table*}

\section{Evaluation Experiments}
\label{sec:Evaluation}

\subsection{Evaluation Settings}
We evaluate all the combinations between four RAG workflows and eight LLMs. However, as FC (Function calling) is implemented by OpenAI, only GPT-3.5 and GPT-4 can use it. Besides, the DFSDT also resembles the format of API calling of FC, therefore, for open-sourced models,  only toolllama that trained on this format can use it. Therefore, we get total 21 feasible RALLM systems and report their results on aminer domain (Table~\ref{tab:aminer_res}) and wiki domain (Table~\ref{tab:comparison_wiki}). All RALLMs are evaluated under the same one-shot setting, with the same decoding hyper-parameters. More details are in Appendix~\ref{sec:eval_detail}.
In the following sections, we will organize our experiment and analysis results with six research questions (RQs).
% \begin{figure}[t]
%     \centering
%         \includegraphics[width=0.5\textwidth]{figures/sns/PAL_gpt-4-1106-preview_histogram.pdf}
%     \caption{Average performance on different levels' tasks and domains for GPT-4  with the PAL workflow.  }
%     \label{fig:task_domain}
% \end{figure}

\begin{figure}[ht]
    \centering
    \subfigure[PAL]{
        \includegraphics[width=0.23\textwidth]{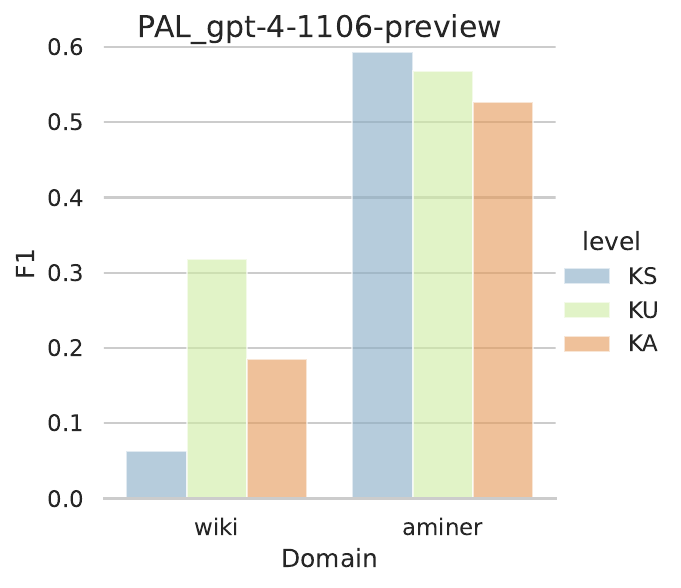}% 设置每张图的宽度为总宽度的四分之一
    }
    \subfigure[DFSDT]{
        \includegraphics[width=0.2\textwidth]{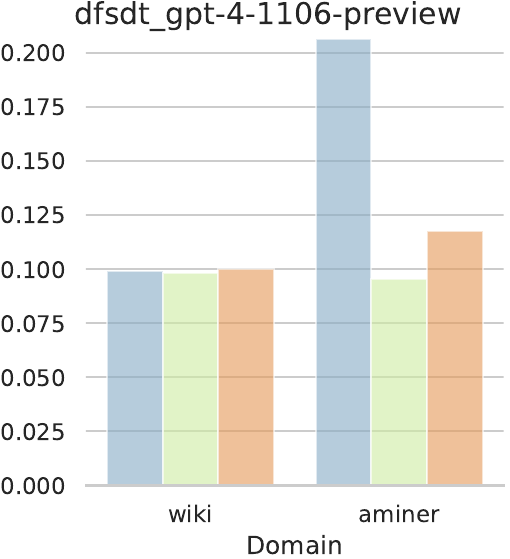}
    }
    \caption{Average performance on different levels' tasks and domains for GPT-4  with the PAL and DFSDT workflow.  }
    \label{fig:task_domain}
\end{figure}

\subsection{Task Results}
\label{sec:Preliminary_EXP}

\textbf{RQ1}: \textit{How effective are RALLMs  across three levels' tasks?}

The evaluation of RALLMs across three levels of tasks, namely Knowledge Seeking (KS), Knowledge Understanding (KU), and Knowledge Application (KA), reveals interesting patterns.

For KS and KA tasks, which assess the model's ability to recall facts and then utilize retrieved knowledge in performing reasoning, the results suggest a potential correlation between a model's rank on KS and KA tasks. Some models, such as the ReAct with GPT-4-1106, demonstrate superior performance on both KS and KA tasks, securing the top rank among all RALLMs, which suggests a strong correlation between retrieving relevant facts and applying retrieved knowledge to new problems. On the other hand, some models like PAL with toolllama2-7b struggle with these tasks, indicating potential areas of improvement in retrieval and reasoning.

% For KS and KA tasks, which evaluate the model's capacity to recall facts and subsequently apply the retrieved knowledge in reasoning, the results suggest a potential correlation between a model's rank on KS and KA tasks. Certain models, such as ReAct with GPT-4-1106, exhibit exceptional performance on both KS and KA tasks, securing the top rank among all RALLMs. This suggests a strong interrelation between the abilities to retrieve facts and to apply this retrieved knowledge to novel problems. Conversely, some models, like PAL with toolllama2-7b, grapple with these tasks, pointing to potential areas for enhancement in retrieval and reasoning capabilities.

In the Knowledge Understanding tasks, which evaluate the model's comprehension of the inherent knowledge embedded in texts, the performance varies as well. Some models that are good at KS and KA tasks perform not as well on KU tasks. For example, ReAct with GPT-4 only ranks the third on aminer's KU task, suggesting that these models may have difficulty interpreting or understanding the underlying knowledge in a given text. However, models with PAL workflow show a strong understanding performance. For example, PAL with GPT-4 performs the best on the KU task of aminer domain. In Figure ~\ref{fig:task_domain}, we observe that this system can also perform well on KU tasks of Wiki domain, proving its ability on understanding.

In summary, the effectiveness of RALLMs across the three levels of tasks varies significantly. While some models excel in certain tasks, they may  falter in  other tasks, highlighting the complexity and challenge of developing systems that can effectively retrieve, understand, and apply domain-specific knowledge.

.
\begin{figure*}[htbp]
    \centering
    \subfigure[ReAct]{
        \includegraphics[width=0.23\textwidth]{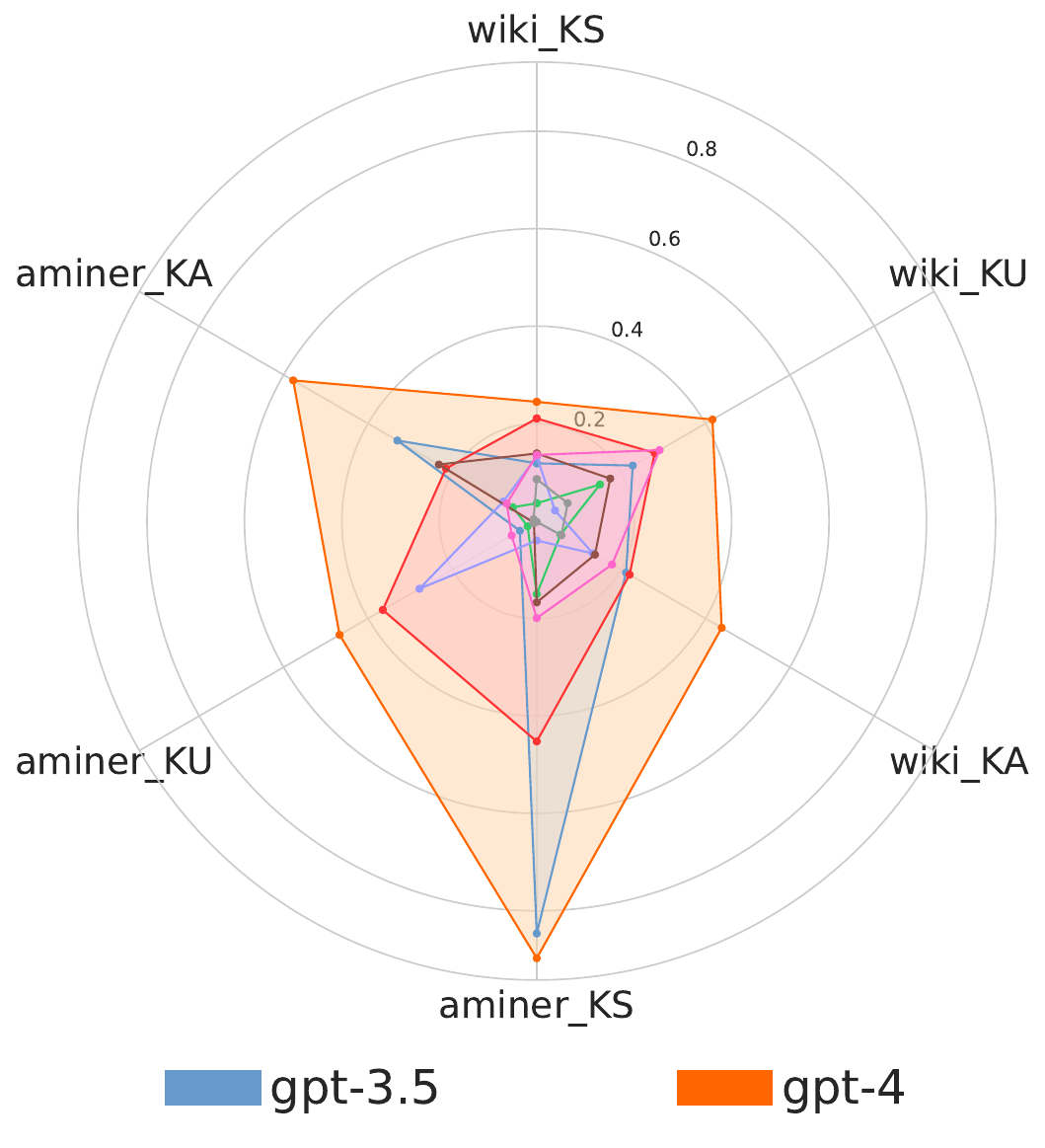}% 设置每张图的宽度为总宽度的四分之一
    }
    \subfigure[PAL]{
        \includegraphics[width=0.23\textwidth]{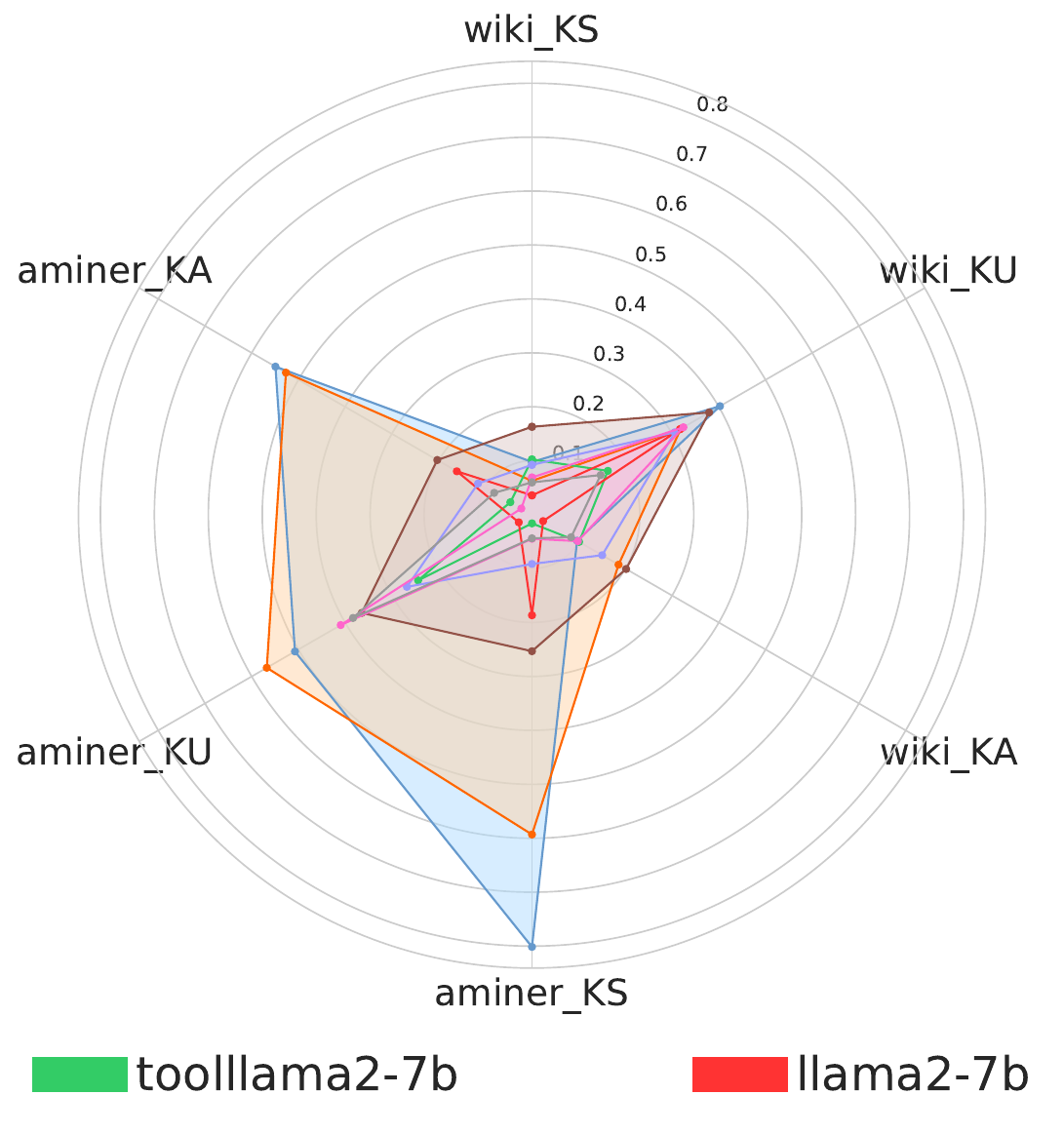}
    }
    \subfigure[DFSDT]{
        \includegraphics[width=0.23\textwidth]{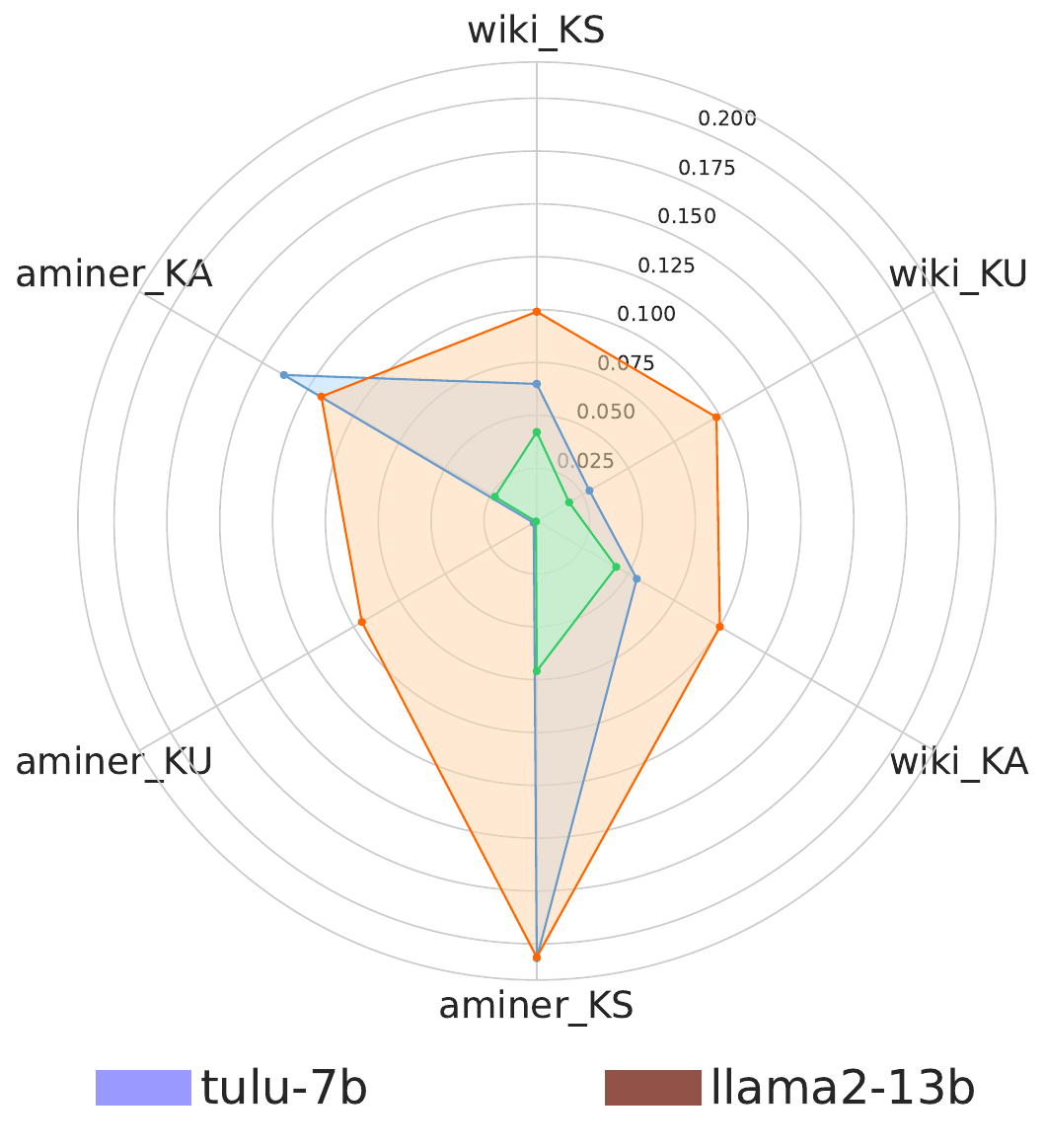}
    }
    \subfigure[FC]{
        \includegraphics[width=0.23\textwidth]{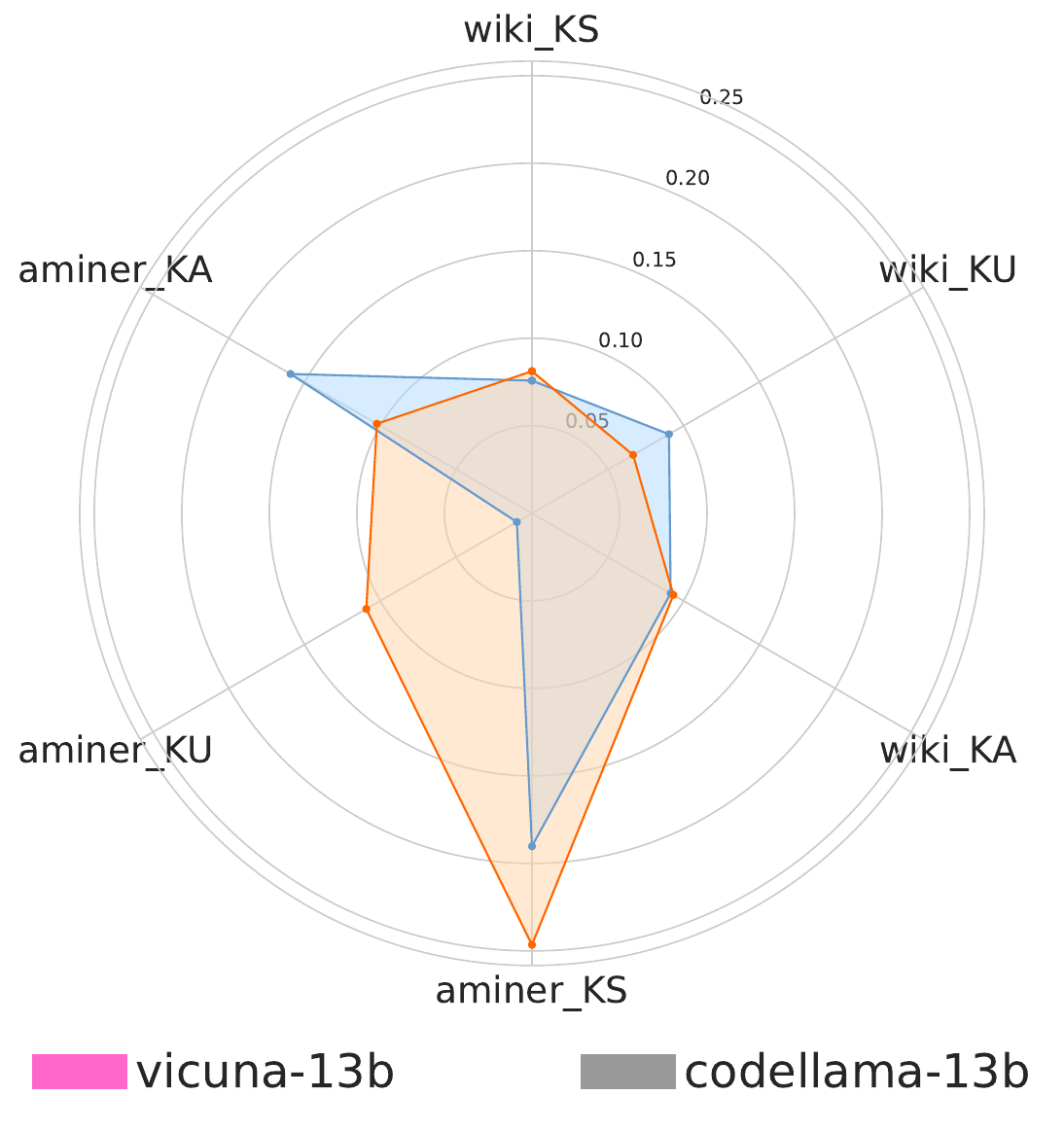}
    }
    \caption{Radar map of single system's performance on all tasks for different $4$ workflows.}
    \label{fig:all_performances}
\end{figure*}
\subsection{Domain Results}

\textbf{RQ2}: \textit{How effective are RALLMs on wiki and aminer domain?}

The effectiveness of RALLMs also varies significantly across different domains. In the Aminer domain, which has over 69 million scholar profiles and 290 million publications, the performance of RALLMs is varied. Some models, like ReAct with GPT-4-1106, perform well, indicating a strong ability to handle domain-specific knowledge. However, other models may encounter challenges within this domain. As illustrated in Table~\ref{tab:aminer_res}, models using the ReAct workflow generally secure higher performance ranks for the Wikipedia domain than for the Aminer domain. This implies that while these models are adept at handling knowledge from a broader domain, they might struggle with the specific types of knowledge and data present within the Aminer domain. %ReAct workflow models generally have a higher performance ranks for Wikipedia domain than their ranks for aminer domain. This suggests that these models are effective at dealing with knowledge from a broader domain, but they may have difficulty dealing with the specific types of knowledge and data present in the Aminer domain.

Overall, while some RALLMs demonstrate strong performance across both the Wikipedia and Aminer domains, others struggle in one or both domains. This highlights the importance of developing models that can effectively handle both broad, open-domain knowledge and domain-specific knowledge.

\subsection{System Comparison}

\textbf{RQ3}: \textit{Which RAG workflow and LLM combination is the best?}

From the results, it appears that the combination of the ReAct workflow with the GPT-4-1106 LLM performs exceptionally well across both the Wikipedia and Aminer domains, as well as across all three levels of tasks. This combination seems to provide a strong balance of fact retrieval, knowledge understanding and application, making it a robust choice for a variety of tasks and domains.

However, it's important to note that while this combination outperforms others in this evaluation, the "best" combination may vary depending on the specific task or domain at hand. For example, other combinations, such as PAL with GPT-4-1106 or ReAct with Llama2-7B-chat, also show strong performance in certain tasks or domains, where PAL with GPT-4-1106 even surpasses ReAct with GPT-4-1106 on the understanding task of aminer domain.

Therefore, while the ReAct and GPT-4-1106 combination appears to be the best overall, other combinations may be more suitable for specific tasks or domains. This underscores the importance of considering both the task and the domain when choosing a RAG workflow and LLM combination.

\section{Analysis}
\label{sec:fg_features}

To discover the underlying characteristics of RALLMs beyond evaluation performance, we delve into a multifaceted analysis on the compatibility between RAG workflows and LLMs, the error types, and the trade-off between effectiveness and efficiency.

\subsection{Matching Analysis}
% \textbf{RQ4}: \textit{What's the best matching LLM for each RAG workflow?}

% To understand the matches between LLMs and RAG workflows better, we conduct an empirical analysis that compares all the LLMs' performances under each RAG workflow. Our visualizations are shown in Figure ~\ref{fig:all_performances}, each sub-figure represents a workflow, which displays all LLMs' average performances on $3$ levels' tasks for $2$ domains. Our findings are follows: (1) On the ReAct workflow~\citep{yao2022react}, gpt-4-1106 is found to be the best matching LLM for all the $6$ kinds of tasks and domains, leading a obvious margin against other LLMs. (2) In terms of the PAL workflow~\citep{gao2023pal}, however, gpt-3.5-turbo-1106 can reach a comparable results with gpt-4-1106 on all levels.  While GPT series models can perform better than other LLMs, they fail to outperform Llama2-13b on the three task levels of wiki domain. This phenomenon reveals that the PAL workflow may boost higher performance for smaller 13b-size LLMs than larger LLMs like GPTs. (3) For the DFSDT workflow~\citep{qin2023toolllm}, as it requires the specific tool using format as FC, only GPTs and toolllama2-7b can generate valid results. It turns out that gpt-4-1106 outperforms the other 2 LLMs on this workflow. (4) For the FC workflow, it is surprising that gpt-4-1106 only beats gpt-3.5-turbo-1106 on KU and KS tasks on aminer domain while it obtains worse or similar results on other tasks. This anomaly inspired us to analyze the error types of gpt-4.

\textbf{RQ4}: \textit{Which LLM best matches each RAG workflow?}

To understand the synergies between LLMs and RAG workflows, we compare the performances of all LLMs within each RAG workflow. The results, visualized in Figure ~\ref{fig:all_performances}, display the average performance of all RALLMs on different tasks, with each sub-figure representing a different workflow. Our findings are as follows:

(1) Within the ReAct workflow~\citep{yao2022react}, GPT-4-1106 emerges as the best matching LLM across all six task types and domains, leading with a significant margin over other LLMs.

(2) However, within the PAL workflow~\citep{gao2023pal}, GPT-3.5-Turbo-1106 achieves results comparable to GPT-4-1106 across all levels. While the GPT series models generally outperform other LLMs, they do not surpass Llama2-13b on the three task levels within the Wikipedia domain. This observation suggests that the PAL workflow may enhance the performance of smaller 13b-size LLMs, such as Llama2-13b, more than larger LLMs like the GPT series.

% (3) Regarding the DFSDT workflow~\citep{qin2023toolllm}, which requires a specific tool-using format akin to FC, only GPT-3.5-Turbo, GPT-4 and ToolLLaMA2-7B~\citep{qin2023toolllm} can generate valid results. Among these, GPT-4-1106 outperforms the other two LLMs by a large margin.

(3) Surprisingly, within the FC workflow, GPT-4 only surpasses GPT-3.5-Turbo on KU and KS tasks within the Aminer domain, while achieving similar or worse results on other tasks. This anomaly prompted us to analyse the error types associated with GPT-4.

\begin{figure*}[htbp]
    \centering
    \subfigure[ReAct]{
        \includegraphics[width=0.23\textwidth]{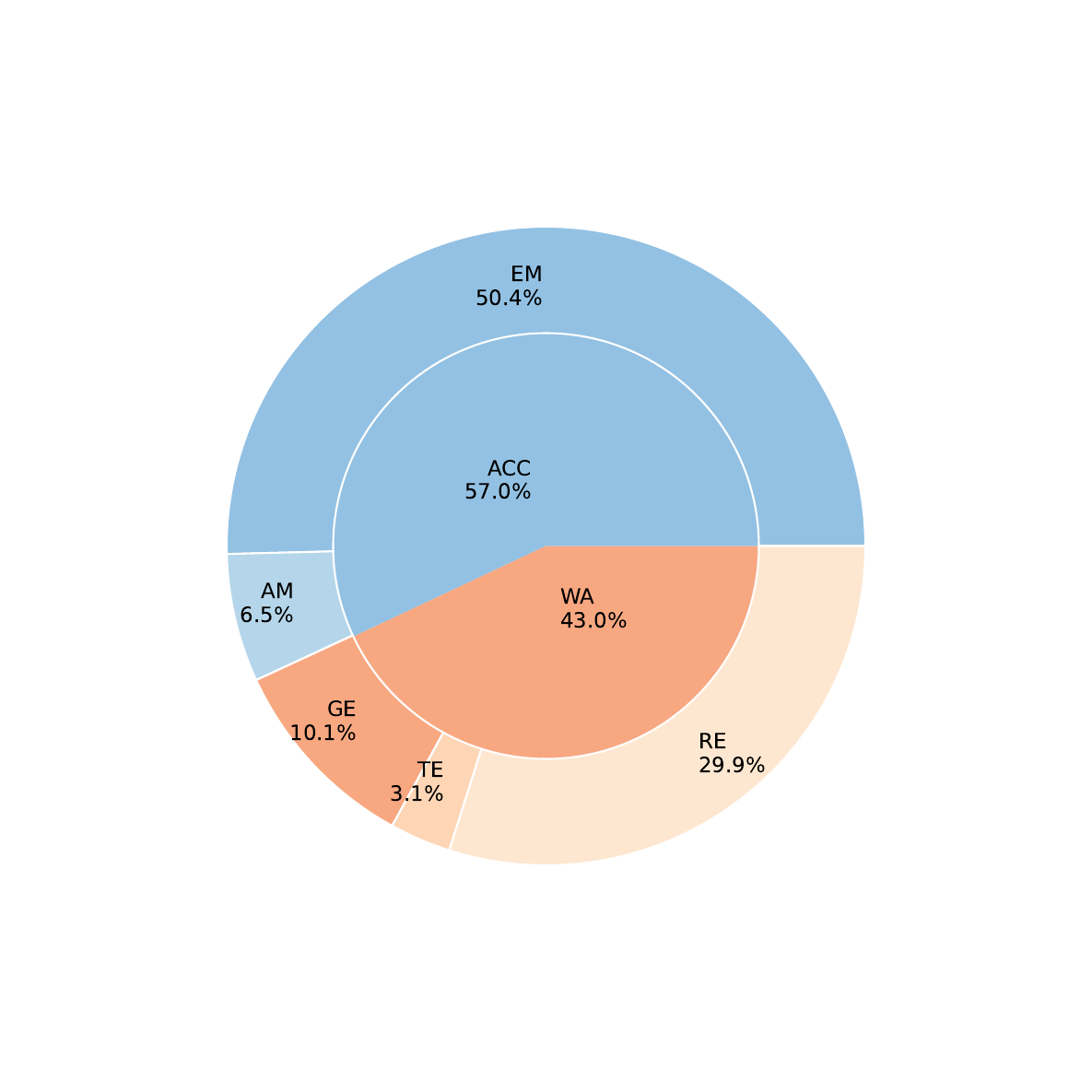}% 设置每张图的宽度为总宽度的四分之一
    }
    \subfigure[PAL]{
        \includegraphics[width=0.23\textwidth]{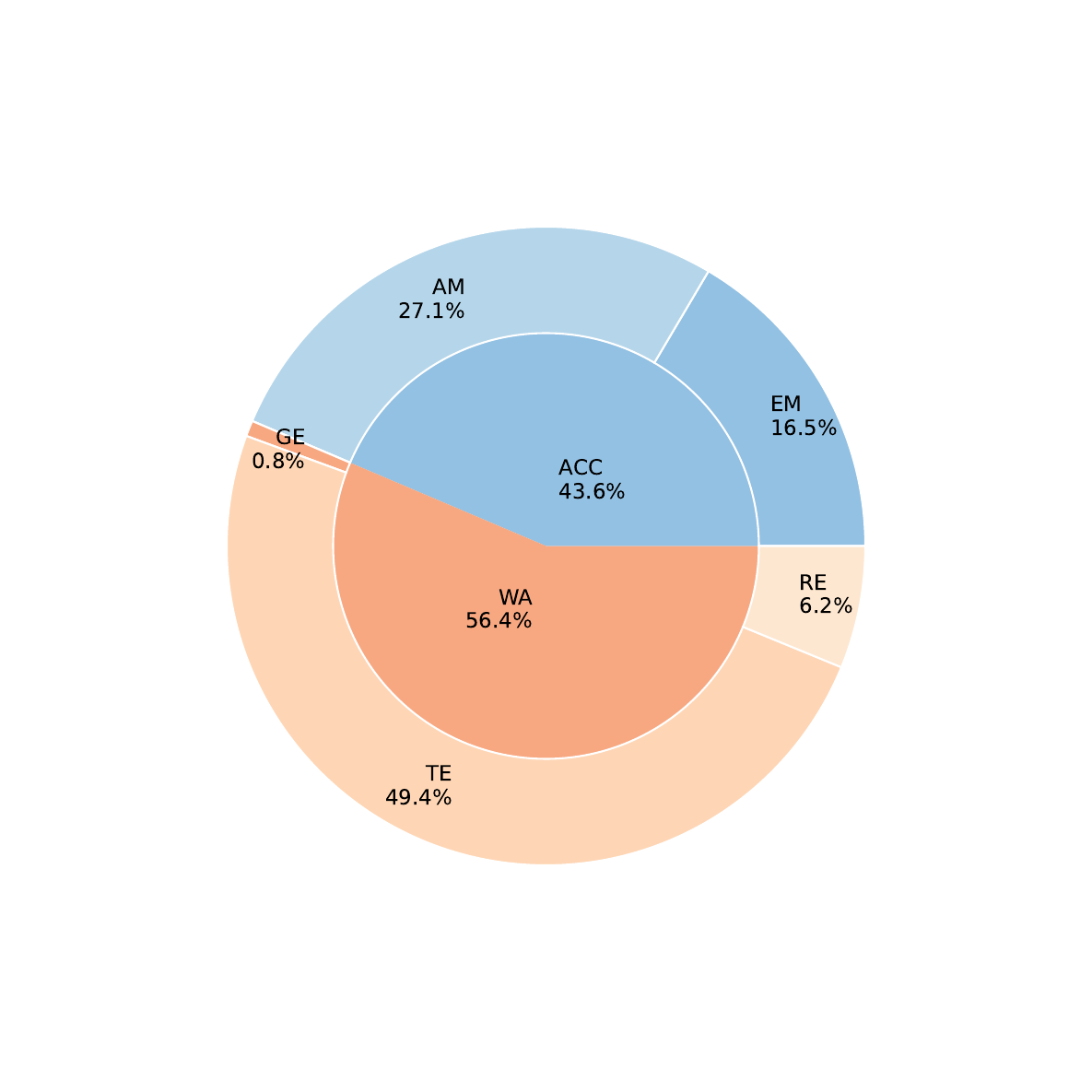}
    }
    \subfigure[DFSDT]{
        \includegraphics[width=0.23\textwidth]{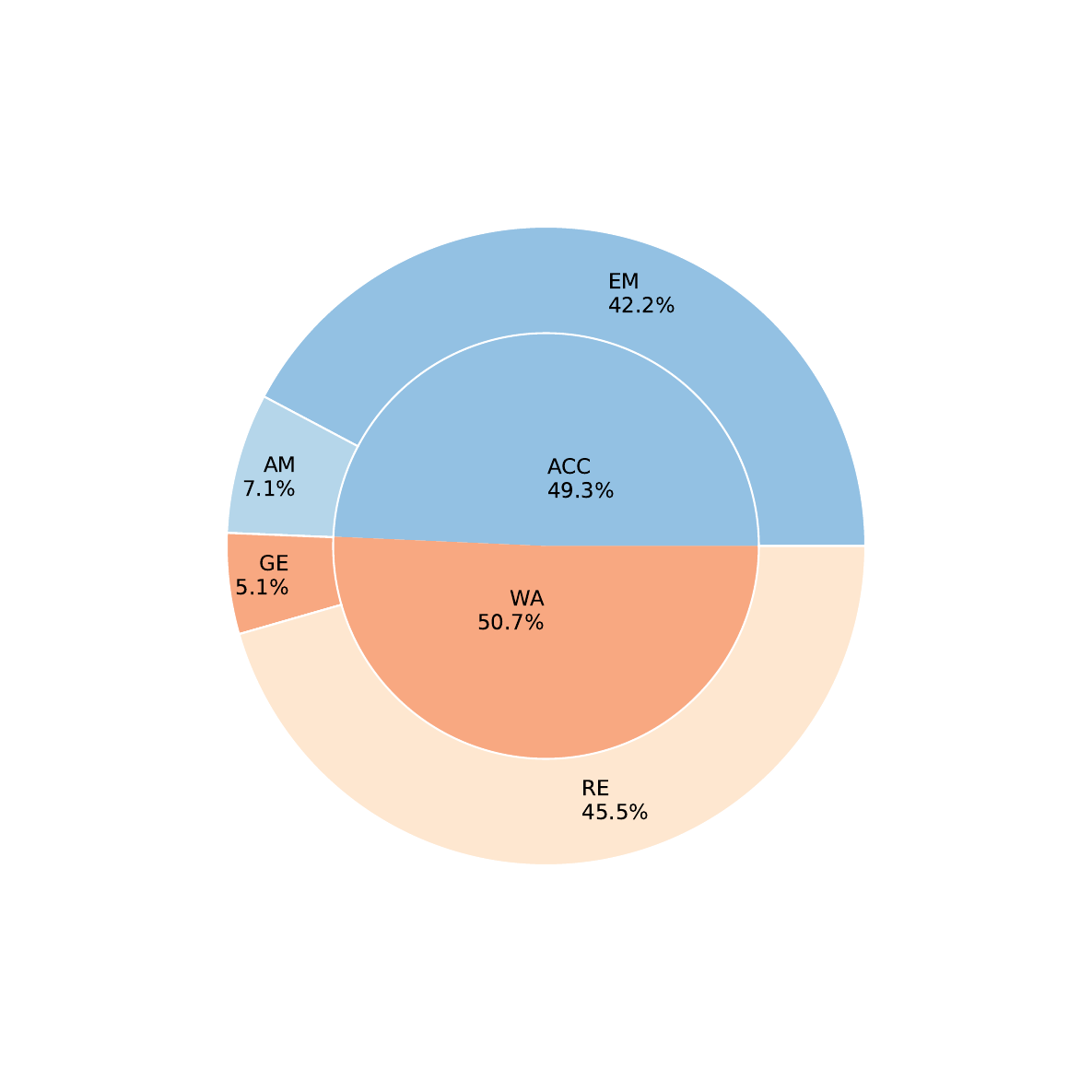}
    }
    \subfigure[FC]{
        \includegraphics[width=0.23\textwidth]{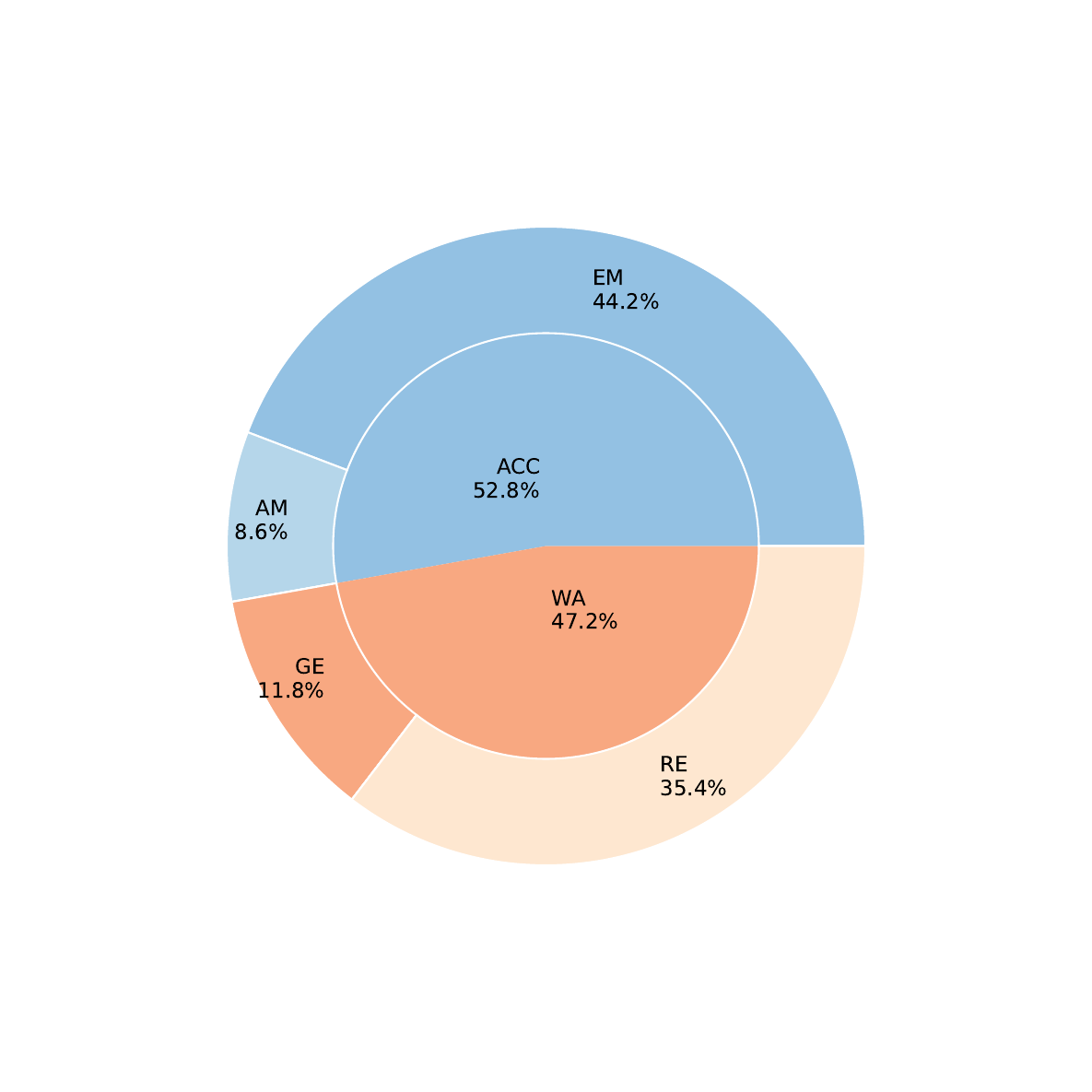}
    }
    \caption{Error distribution of GPT-4-preview-1104 with different $4$ workflows on all tasks. }
    \label{fig:all_error}
\end{figure*}

\begin{figure}[t]
    \centering
        \includegraphics[width=0.5\textwidth]{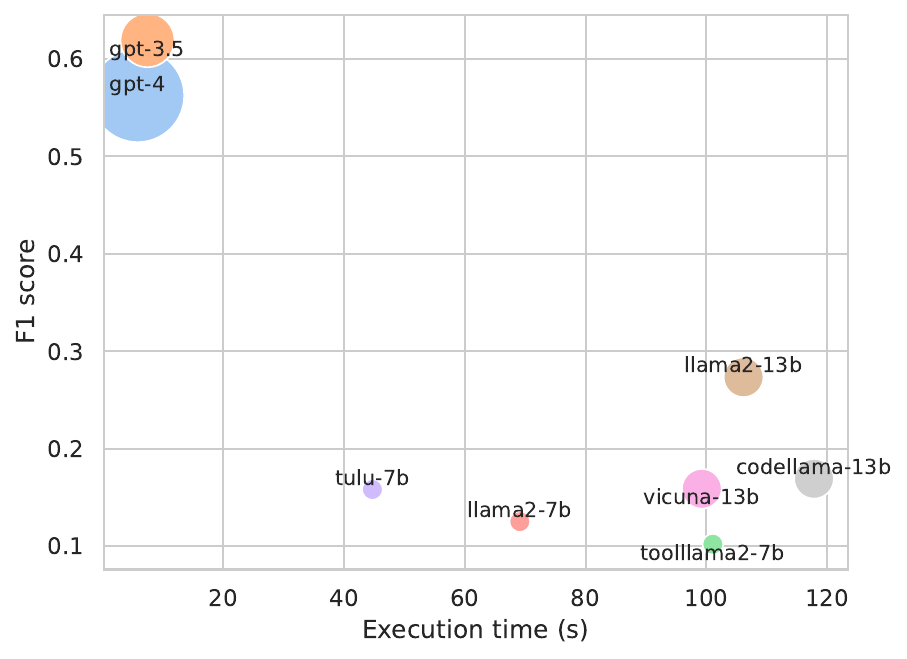}
    \caption{Average time cost and task performance of different LLMs with PAL workflow on aminer domain for deployment.  }
    \label{fig:deploy}
\end{figure}

\subsection{Error Analysis}

% \textbf{RQ5}: \textit{What kinds of errors does gpt-4 make on different workflows?}

% Previously in Section ~\ref{analysis_tools}, we have defined the error types of RALLM systems. In this section, we attempt to identify the proportion of gpt-4's error types on different RAG workflows.

% In Figure ~\ref{fig:all_error}, we visualize the error type distribution of gpt-4 with the $4$ RAG workflows. We find that the PAL workflow has the highest Answer Match portion among all workflows, which is 27.1\% while its Exact Match answers only takes 16.5\%. This means knowledge retrieved by PAL didn't contain much useful information for reasoning, most right answers comes from LLMs' inner knowledge. Besides, the tool-using error for PAL is also the highest, reaching 49.4\%, which means that the API calling process is usually halted so that there is no domain knowledge can be successfully retrieved. This provides a crucial insight about the behaviour of PAL, which explains why gpt-4 performs bad on PAL as it only execute the API calling code once so that it can not revise the code when confronted with a tool execution error.

% For other RAG workflows, their tool using manners are more robust but cost more inference time as they can interact with our query APIs to get feedback and can query for multiple rounds. Therefore, the ReAct, DFSDT and FC workflows produce less tool using error than PAL. So they can more successfully retrieve the domain knowledge to help them solve the problems, resulting in less AM and more EM numbers for the answers.

\textbf{RQ5}: \textit{What types of errors does GPT-4 make across different workflows?}

% \begin Shangqing 原版
In Section ~\ref{analysis_tools}, we defined various error types that RALLM systems might encounter. Here, we aim to identify the distribution of these error types in GPT-4 across different RAG workflows.

Figure ~\ref{fig:all_error} visualizes the distribution of GPT-4's error types across the four RAG workflows. Notably, the PAL workflow exhibits the highest proportion of Answer Match errors among all workflows at 27.1\%, while its Exact Match answers only account for 16.5\%. This suggests that the knowledge retrieved by PAL does not contribute significantly to reasoning, and most correct answers originate from the inherent knowledge of the LLMs. Additionally, the tool-using error rate for PAL is also the highest, reaching 49.4\%, indicating that the API calling process is often interrupted, preventing successful retrieval of domain knowledge. As PAL workflow executes the API calling code only once, it cannot revise the code when encountering a tool execution error, which is critical in other workflows~\citep{yao2022react,toolllm} and explains why GPT-4 performs poorly with PAL. 

Other RAG workflows, while more robust in their tool-using manners, require more inference time as they can interact with our query APIs for feedback and can execute multiple rounds of queries. Consequently, the ReAct, DFSDT, and FC workflows yield fewer tool-using errors than PAL, enabling them to retrieve domain knowledge more successfully. This results in fewer Answer Match errors and a greater number of Exact Match answers.

\subsection{Deployment Analysis}
\label{application}

% \textbf{RQ6}: \textit{Which system provides the best practical performance (efficiency and effectiveness) on wiki and aminer domain?}

% We are interested in evaluating the practical performance of these RALLM systems in terms of execution time and answer F1 score. We report the results of 8 LLMs with PAL workflow for this study, since PAL has the simple-turn interaction with the environment, making it much faster than other workflow. Following ~\cite{toolllm}, we conduct efficiency evaluation using one single thread on a GPU (Nvidia A100.) for each open-source LLM, and one single thread for each OpenAI api-based LLM. The analysis result for aminer domain is shown in Figure ~\ref{fig:deploy}, other results are shown in Appendix ~\ref{app:deploy}. GPT-4 and GPT-3.5 get better F1 scores on Aminer and cost less execution time than other open-source LLMs. That's because we only run each open-source LLM on single GPU while GPT models are called through API where they may have a cloud server with many GPUs.

% If we only compare open-source LLMs, we find that although llama-13b is about the best on F1 score, it is much slower other LLMs (105s per query on the aminer domain). Among all the models, tulu-7b achieves a good trade-off between efficiency (40-50s per query) and effectiveness (F1 ~0.18) on the aminer domain. Interestingly from Figure ~\ref{fig:deploy}, we find the open-source LLMs with larger parameters can significantly increase the F1 score while decrease on efficiency. This trade-off can help developers choosing the suitable LLM for their domain applications.

\textbf{RQ6}: \textit{Which system offers the best practical performance (both in terms of efficiency and effectiveness) within the specific domain?}

We are interested in assessing the practical performance of these RALLM systems, specifically in terms of execution time and the F1 score of the answers. For this study, we report the results of eight LLMs using the PAL workflow, as PAL's simple-turn interaction with the environment allows for faster execution compared to other workflows. Following the evaluation process of ToolBench~\cite{toolllm}, we evaluate efficiency using a single thread on a GPU (Nvidia A100) for each open-source LLM, and one single thread for each LLM called via the OpenAI API. The analysis results for the Aminer domain are illustrated in Figure ~\ref{fig:deploy}, with additional results provided in Appendix ~\ref{app:deploy}. Our findings are as follows:

GPT-4 and GPT-3.5 achieve superior F1 scores on Aminer and require less execution time than other open-source LLMs. This efficiency is likely due to the fact that while each open-source LLM is run on a single GPU, the GPT models are invoked through an API, which may leverage a cloud server with multiple GPUs.

When comparing only open-source LLMs, we observe that although Llama-13b ranks among the best in terms of F1 score, it is significantly slower than other LLMs, taking approximately 105 seconds per query in the Aminer domain. Among all the models, Tulu-7b strikes a commendable balance between efficiency (40-50 seconds per query) and effectiveness (F1 ~0.18). Intriguingly, as depicted in Figure ~\ref{fig:deploy}, we find that open-source LLMs with larger parameters can markedly improve the F1 score while compromising on efficiency. This trade-off can guide developers in selecting the most suitable RALLM for their specific domain applications.

\section{Conclusion}

In this work, we presented the R-Eval toolkit, a comprehensive evaluation platform designed to address the existing gap in the systematic evaluation of Retrieval-Augmented Large Language Models (RALLMs). By offering a user-friendly, modular, and extensible interface, R-Eval facilitates the comparison and analysis of various RAG workflows and LLMs. Our study using this toolkit revealed significant variations in the performance of RALLMs across different tasks and domains, underscoring the importance of task- and domain-specific selection of RAG workflows and LLMs. As the field continues to evolve, we believe that R-Eval will 
 benefit both researchers and industry professionals with a pivotal role in shaping the future of RALLMs and their domain-specific applications.

\section*{Acknowledgement}
This work is supported by the National Key Research \& Develop Plan (2023YFF0725100), a grant from the Institute for Guo Qiang, Tsinghua University (2019GQB0003), Tsinghua University Initiative Scientific Research Program and Zhipu AI. 
Jing Zhang is supported by
the NSF of China (62322214).

\bibliographystyle{ACM-Reference-Format}
\bibliography{sample-base}

\appendix

\section{Implementation Details}
\label{sec:implementation}

We have implemented an environment for each domain with several query APIs to retrieve the domain knowledge.  

On Aminer domain, we have these APIs:

$\bullet$ \textbf{searchPerson}. The searchPerson function, which is based on the scholar entities' information in Aminer, receives the name, organization and interest of this intended scholar and returns the detailed information including person's id, citation number and publication number  via fuzzy match.

$\bullet$ \textbf{searchPublication}. The searchPublication function, which is based on the publication entities' information in Aminer, receives the publication information and returns the related information including publication's id, title and publication year via fuzzy match.

$\bullet$ \textbf{getCoauthors}. The getCoauthors function, which is based on the relation information between scholar entities, receives the person id then returns the input scholar's coauthors and their detailed information including id, name and relation via exact match.

$\bullet$ \textbf{getPersonInterest}. The getPersonInterest function, which is based on the property information of scholar entities in Aminer, receives the scholar's id and returns a list of the person's interested research topics via exact match.

$\bullet$ \textbf{getPublication}. The getPublication function, which is based on the property information of publication entities in Aminer, receives the publication's id and returns its detailed information including the publication's abstract, author list and the number of citation via exact match.

$\bullet$ \textbf{getPersonBasicInfo}. The getPersonBasicInfo function, which is based on the scholar entities' property information in Aminer, receives the person's id of this intended scholar and returns the detailed information including person's name, gender, organization, position, short bio, education experience and email address via exact match. In fact, these information consists of the person's profile.

$\bullet$ \textbf{getPersonPubs}. The getPersonPubs function, which is based on the relation information between publication entities and  scholar entities in Aminer, receives the person's id, and returns the detailed information including the publication's id, title, citation number and the authors' name list via exact match.

On Wikipedia domain, we have these APIs:

$\bullet$ \textbf{Search}. The Search function, which is based on the entities' page information in Wikipedia, receives the entity's name, and returns the page's abstract via fuzzy match while storing the other sections' information in the document store for the further usage. If there is no matching Wikipedia entity, the API will provide a list of possibly related entities for continuous searching.

$\bullet$ \textbf{Lookup}. The Lookup function, which is based on the previous stored entity information via search API on Wikipedia, receives the keyword and returns a list of  relevant text segments from the document store via fuzzy match.

$\bullet$ \textbf{Finish}. This is a special function for stopping the searching process on Wikipedia.

\begin{table}[ht]
      \caption{Hyper-parameters for open-source LLMs' inference in our evaluation.  }
    \centering
    %\small
    \begin{tabular}{c|c|c}
        \toprule
\textbf{Process}   &   \textbf{Parameter} & \textbf{Value}   \\
                \midrule 
         
  \multirow{3}{*}{ \shortstack{Tokenization} }  
  & max\_length  & 2048   \\
  & truncation  & True   \\
  & skip\_special\_tokens  & True   \\
         \midrule 

 \multirow{5}{*}{ \shortstack{Decoding} } %& XDAI & 36.7 & 14.3 & 33.8  \\
  &num\_beams  & 1  \\
  & do\_sample   & False   \\
  & temperature  & 0   \\
        & stop sequence  & </s> \\
   
   & max\_new\_tokens   & 128  \\
        \bottomrule
    \end{tabular}
         
    \label{tab:hyper_parameters}
\end{table}

\begin{table*}[t]
\caption{Performance of different systems for each task on Wikipedia domain.}
\centering
  % \small
% \resizebox{\linewidth}{!}{
\begin{tabular}{l|l|rrr|rrrr|rrrrr}
\toprule
\multirow{2}{*}{\textbf{Workflow}} & \multirow{2}{*}{\textbf{LLM}} & \multicolumn{3}{c}{\textbf{\cellcolor{'shallow1'} wiki KS}}                                                                                                                                                                                                            & \multicolumn{4}{|c}{\textbf{\cellcolor{'shallow2'} wiki KU}}     
                                                                 & \multicolumn{5}{|c}{\textbf{\cellcolor{'shallow3'} wiki KA}}    \\ 
                             &   & \multicolumn{1}{c}{\cellcolor{'deep1'} \textbf{1-1}}  & \multicolumn{1}{c}{\cellcolor{'deep1'} \textbf{1-2}}  & \multicolumn{1}{c}{\cellcolor{'deep1'} \textbf{Rank}} & \multicolumn{1}{|c}{\cellcolor{'deep2'} \textbf{2-1}} & \multicolumn{1}{c}{\cellcolor{'deep2'} \textbf{2-2}}  & \multicolumn{1}{c}{\cellcolor{'deep2'} \textbf{2-3}}  & \multicolumn{1}{c}{\cellcolor{'deep2'} \textbf{Rank}} & \multicolumn{1}{|c}{\cellcolor{'deep3'} \textbf{3-1}} & \multicolumn{1}{c}{\cellcolor{'deep3'} \textbf{3-2}} & \multicolumn{1}{c}{\cellcolor{'deep3'} \textbf{3-3}} & \multicolumn{1}{c}{\cellcolor{'deep3'} \textbf{3-4}} & \multicolumn{1}{c}{\cellcolor{'deep3'} \textbf{Rank}}  \\ \midrule
\cellcolor{'wit'} ReAct & \cellcolor{'wit'} gpt-4-1106 & \cellcolor{'shallow1'} 23.1 & \cellcolor{'shallow1'} 25.8 & \cellcolor{'shallow1'}  1st & \cellcolor{'shallow2'} 33.7 & \cellcolor{'shallow2'} 51.3 & \cellcolor{'shallow2'} 40.0 & \cellcolor{'shallow2'}  1st & \cellcolor{'shallow3'} 59.7 & \cellcolor{'shallow3'} 64.6 & \cellcolor{'shallow3'} 27.3 & \cellcolor{'shallow3'} 23.8 & \cellcolor{'shallow3'}  1st  \\ 
\cellcolor{'gry'} PAL & \cellcolor{'gry'} gpt-3.5-turbo & \cellcolor{'deep1'} 10.4 & \cellcolor{'deep1'} 9.1 & \cellcolor{'deep1'}  10th & \cellcolor{'deep2'} 12.7 & \cellcolor{'deep2'} 65.0 & \cellcolor{'deep2'} 43.0 & \cellcolor{'deep2'}  2nd & \cellcolor{'deep3'} 12.1 & \cellcolor{'deep3'} 9.4 & \cellcolor{'deep3'} 5.1 & \cellcolor{'deep3'} 12.0 & \cellcolor{'deep3'}  13th  \\ 
\cellcolor{'wit'} PAL & \cellcolor{'wit'} gpt-4-1106 & \cellcolor{'shallow1'} 7.9 & \cellcolor{'shallow1'} 4.6 & \cellcolor{'shallow1'}  17th & \cellcolor{'shallow2'} 18.6 & \cellcolor{'shallow2'} 30.0 & \cellcolor{'shallow2'} 47.0 & \cellcolor{'shallow2'}  5th & \cellcolor{'shallow3'} 30.3 & \cellcolor{'shallow3'} 12.1 & \cellcolor{'shallow3'} 14.2 & \cellcolor{'shallow3'} 17.5 & \cellcolor{'shallow3'}  5th  \\ 
\cellcolor{'gry'} ReAct & \cellcolor{'gry'} llama2-7b-chat & \cellcolor{'deep1'} 21.2 & \cellcolor{'deep1'} 20.9 & \cellcolor{'deep1'}  2nd & \cellcolor{'deep2'} 2.0 & \cellcolor{'deep2'} 55.0 & \cellcolor{'deep2'} 26.9 & \cellcolor{'deep2'}  9th & \cellcolor{'deep3'} 31.5 & \cellcolor{'deep3'} 26.8 & \cellcolor{'deep3'} 11.5 & \cellcolor{'deep3'} 18.3 & \cellcolor{'deep3'}  2nd  \\ 
\cellcolor{'wit'} PAL & \cellcolor{'wit'} llama2-13b & \cellcolor{'shallow1'} 19.1 & \cellcolor{'shallow1'} 13.5 & \cellcolor{'shallow1'}  3rd & \cellcolor{'shallow2'} 1.3 & \cellcolor{'shallow2'} 67.0 & \cellcolor{'shallow2'} 45.4 & \cellcolor{'shallow2'}  3rd & \cellcolor{'shallow3'} 25.0 & \cellcolor{'shallow3'} 18.5 & \cellcolor{'shallow3'} 8.0 & \cellcolor{'shallow3'} 29.0 & \cellcolor{'shallow3'}  4th  \\ 
\cellcolor{'gry'} ReAct & \cellcolor{'gry'} gpt-3.5-turbo & \cellcolor{'deep1'} 11.2 & \cellcolor{'deep1'} 12.5 & \cellcolor{'deep1'}  7th & \cellcolor{'deep2'} 9.3 & \cellcolor{'deep2'} 27.0 & \cellcolor{'deep2'} 32.0 & \cellcolor{'deep2'}  10th & \cellcolor{'deep3'} 43.1 & \cellcolor{'deep3'} 30.1 & \cellcolor{'deep3'} 7.6 & \cellcolor{'deep3'} 3.8 & \cellcolor{'deep3'}  3rd  \\ 
\cellcolor{'wit'} ReAct & \cellcolor{'wit'} vicuna-13b & \cellcolor{'shallow1'} 14.9 & \cellcolor{'shallow1'} 12.3 & \cellcolor{'shallow1'}  5th & \cellcolor{'shallow2'} 4.3 & \cellcolor{'shallow2'} 57.1 & \cellcolor{'shallow2'} 26.0 & \cellcolor{'shallow2'}  8th & \cellcolor{'shallow3'} 25.2 & \cellcolor{'shallow3'} 28.0 & \cellcolor{'shallow3'} 4.3 & \cellcolor{'shallow3'} 14.0 & \cellcolor{'shallow3'}  6th  \\ 
\cellcolor{'gry'} PAL & \cellcolor{'gry'} tulu-7b & \cellcolor{'deep1'} 10.4 & \cellcolor{'deep1'} 8.1 & \cellcolor{'deep1'}  11th & \cellcolor{'deep2'} 2.7 & \cellcolor{'deep2'} 34.8 & \cellcolor{'deep2'} 54.2 & \cellcolor{'deep2'}  7th & \cellcolor{'deep3'} 15.1 & \cellcolor{'deep3'} 22.7 & \cellcolor{'deep3'} 7.5 & \cellcolor{'deep3'} 14.8 & \cellcolor{'deep3'}  7th  \\ 
\cellcolor{'wit'} PAL & \cellcolor{'wit'} vicuna-13b & \cellcolor{'shallow1'} 8.4 & \cellcolor{'shallow1'} 5.4 & \cellcolor{'shallow1'}  15th & \cellcolor{'shallow2'} 1.0 & \cellcolor{'shallow2'} 49.9 & \cellcolor{'shallow2'} 46.4 & \cellcolor{'shallow2'}  4th & \cellcolor{'shallow3'} 11.9 & \cellcolor{'shallow3'} 11.2 & \cellcolor{'shallow3'} 2.9 & \cellcolor{'shallow3'} 13.3 & \cellcolor{'shallow3'}  12th  \\ 
\cellcolor{'gry'} ReAct & \cellcolor{'gry'} llama2-13b & \cellcolor{'deep1'} 16.2 & \cellcolor{'deep1'} 11.5 & \cellcolor{'deep1'}  4th & \cellcolor{'deep2'} 0.2 & \cellcolor{'deep2'} 30.0 & \cellcolor{'deep2'} 22.0 & \cellcolor{'deep2'}  11th & \cellcolor{'deep3'} 20.1 & \cellcolor{'deep3'} 24.6 & \cellcolor{'deep3'} 4.0 & \cellcolor{'deep3'} 6.5 & \cellcolor{'deep3'}  8th  \\ 
\cellcolor{'wit'} PAL & \cellcolor{'wit'} llama2-7b-chat & \cellcolor{'shallow1'} 4.0 & \cellcolor{'shallow1'} 3.2 & \cellcolor{'shallow1'}  21th & \cellcolor{'shallow2'} 1.3 & \cellcolor{'shallow2'} 60.3 & \cellcolor{'shallow2'} 33.4 & \cellcolor{'shallow2'}  6th & \cellcolor{'shallow3'} 2.9 & \cellcolor{'shallow3'} 2.4 & \cellcolor{'shallow3'} 0.9 & \cellcolor{'shallow3'} 3.3 & \cellcolor{'shallow3'}  21th  \\ 
\cellcolor{'gry'} PAL & \cellcolor{'gry'} codellama-13b & \cellcolor{'deep1'} 6.0 & \cellcolor{'deep1'} 6.0 & \cellcolor{'deep1'}  18th & \cellcolor{'deep2'} 0.4 & \cellcolor{'deep2'} 11.7 & \cellcolor{'deep2'} 32.0 & \cellcolor{'deep2'}  14th & \cellcolor{'deep3'} 6.6 & \cellcolor{'deep3'} 12.9 & \cellcolor{'deep3'} 4.5 & \cellcolor{'deep3'} 9.5 & \cellcolor{'deep3'}  16th  \\ 
\cellcolor{'wit'} PAL & \cellcolor{'wit'} toolllama2-7b & \cellcolor{'shallow1'} 9.9 & \cellcolor{'shallow1'} 10.6 & \cellcolor{'shallow1'}  8th & \cellcolor{'shallow2'} 2.2 & \cellcolor{'shallow2'} 40.5 & \cellcolor{'shallow2'} 6.1 & \cellcolor{'shallow2'}  12th & \cellcolor{'shallow3'} 8.5 & \cellcolor{'shallow3'} 19.5 & \cellcolor{'shallow3'} 3.5 & \cellcolor{'shallow3'} 8.9 & \cellcolor{'shallow3'}  10th  \\ 
\cellcolor{'gry'} ReAct & \cellcolor{'gry'} tulu-7b & \cellcolor{'deep1'} 14.7 & \cellcolor{'deep1'} 11.8 & \cellcolor{'deep1'}  6th & \cellcolor{'deep2'} 2.0 & \cellcolor{'deep2'} 5.0 & \cellcolor{'deep2'} 6.0 & \cellcolor{'deep2'}  19th & \cellcolor{'deep3'} 18.0 & \cellcolor{'deep3'} 23.3 & \cellcolor{'deep3'} 1.4 & \cellcolor{'deep3'} 10.5 & \cellcolor{'deep3'}  9th  \\ 
\cellcolor{'wit'} DFSDT & \cellcolor{'wit'} gpt-4-1106 & \cellcolor{'shallow1'} 10.8 & \cellcolor{'shallow1'} 9.0 & \cellcolor{'shallow1'}  9th & \cellcolor{'shallow2'} 9.0 & \cellcolor{'shallow2'} 8.2 & \cellcolor{'shallow2'} 12.2 & \cellcolor{'shallow2'}  15th & \cellcolor{'shallow3'} 10.7 & \cellcolor{'shallow3'} 18.4 & \cellcolor{'shallow3'} 7.9 & \cellcolor{'shallow3'} 3.1 & \cellcolor{'shallow3'}  11th  \\ 
\cellcolor{'gry'} FC & \cellcolor{'gry'} gpt-4-1106 & \cellcolor{'deep1'} 8.2 & \cellcolor{'deep1'} 8.0 & \cellcolor{'deep1'}  13th & \cellcolor{'deep2'} 9.9 & \cellcolor{'deep2'} 3.3 & \cellcolor{'deep2'} 6.8 & \cellcolor{'deep2'}  18th & \cellcolor{'deep3'} 10.9 & \cellcolor{'deep3'} 15.0 & \cellcolor{'deep3'} 8.1 & \cellcolor{'deep3'} 3.3 & \cellcolor{'deep3'}  14th  \\ 
\cellcolor{'wit'} FC & \cellcolor{'wit'} gpt-3.5-turbo & \cellcolor{'shallow1'} 8.0 & \cellcolor{'shallow1'} 7.2 & \cellcolor{'shallow1'}  14th & \cellcolor{'shallow2'} 2.1 & \cellcolor{'shallow2'} 4.0 & \cellcolor{'shallow2'} 21.0 & \cellcolor{'shallow2'}  16th & \cellcolor{'shallow3'} 12.4 & \cellcolor{'shallow3'} 15.1 & \cellcolor{'shallow3'} 5.7 & \cellcolor{'shallow3'} 3.4 & \cellcolor{'shallow3'}  15th  \\ 
\cellcolor{'gry'} ReAct & \cellcolor{'gry'} toolllama2-7b & \cellcolor{'deep1'} 5.8 & \cellcolor{'deep1'} 1.5 & \cellcolor{'deep1'}  20th & \cellcolor{'deep2'} 0.0 & \cellcolor{'deep2'} 44.0 & \cellcolor{'deep2'} 1.0 & \cellcolor{'deep2'}  13th & \cellcolor{'deep3'} 4.4 & \cellcolor{'deep3'} 12.1 & \cellcolor{'deep3'} 0.0 & \cellcolor{'deep3'} 6.0 & \cellcolor{'deep3'}  18th  \\ 
\cellcolor{'wit'} DFSDT & \cellcolor{'wit'} gpt-3.5-turbo & \cellcolor{'shallow1'} 6.8 & \cellcolor{'shallow1'} 6.2 & \cellcolor{'shallow1'}  16th & \cellcolor{'shallow2'} 0.9 & \cellcolor{'shallow2'} 4.0 & \cellcolor{'shallow2'} 3.7 & \cellcolor{'shallow2'}  20th & \cellcolor{'shallow3'} 7.6 & \cellcolor{'shallow3'} 10.1 & \cellcolor{'shallow3'} 2.9 & \cellcolor{'shallow3'} 1.3 & \cellcolor{'shallow3'}  19th  \\ 
\cellcolor{'gry'} ReAct & \cellcolor{'gry'} codellama-13b & \cellcolor{'deep1'} 7.7 & \cellcolor{'deep1'} 9.5 & \cellcolor{'deep1'}  12th & \cellcolor{'deep2'} 0.0 & \cellcolor{'deep2'} 13.0 & \cellcolor{'deep2'} 9.0 & \cellcolor{'deep2'}  17th & \cellcolor{'deep3'} 5.7 & \cellcolor{'deep3'} 9.0 & \cellcolor{'deep3'} 0.0 & \cellcolor{'deep3'} 8.7 & \cellcolor{'deep3'}  17th  \\ 
\cellcolor{'wit'} DFSDT & \cellcolor{'wit'} toolllama2-7b & \cellcolor{'shallow1'} 4.5 & \cellcolor{'shallow1'} 4.0 & \cellcolor{'shallow1'}  19th & \cellcolor{'shallow2'} 0.9 & \cellcolor{'shallow2'} 4.1 & \cellcolor{'shallow2'} 0.3 & \cellcolor{'shallow2'}  21th & \cellcolor{'shallow3'} 1.8 & \cellcolor{'shallow3'} 8.7 & \cellcolor{'shallow3'} 2.1 & \cellcolor{'shallow3'} 4.8 & \cellcolor{'shallow3'}  20th  \\ 
\bottomrule
\end{tabular}
\label{tab:comparison_wiki}
\end{table*}

\begin{figure*}[ht]
    \centering
    \subfigure[ReAct]{
        \includegraphics[width=0.33\textwidth]{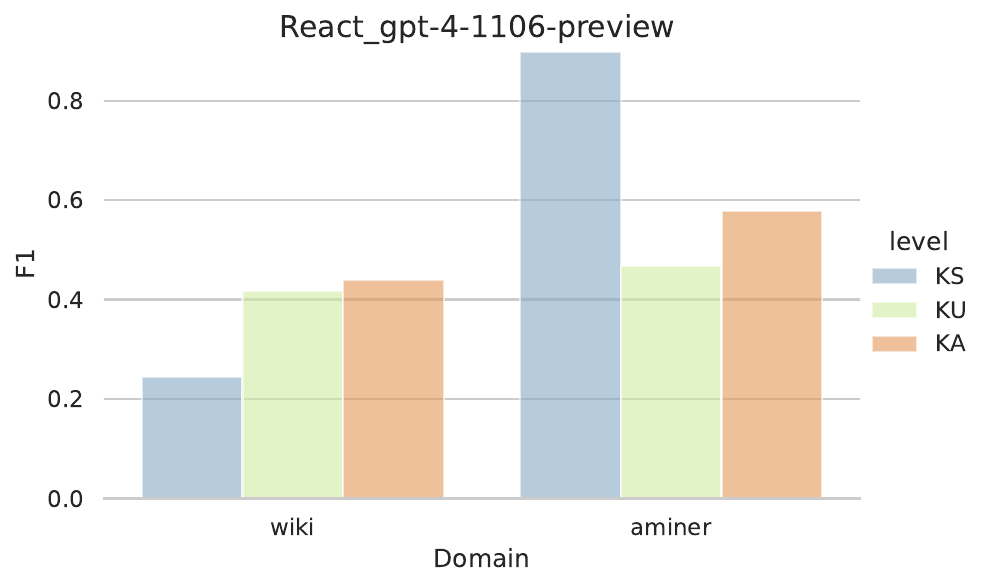}% 设置每张图的宽度为总宽度的四分之一
    }
    \subfigure[DFSDT]{
        \includegraphics[width=0.3\textwidth]{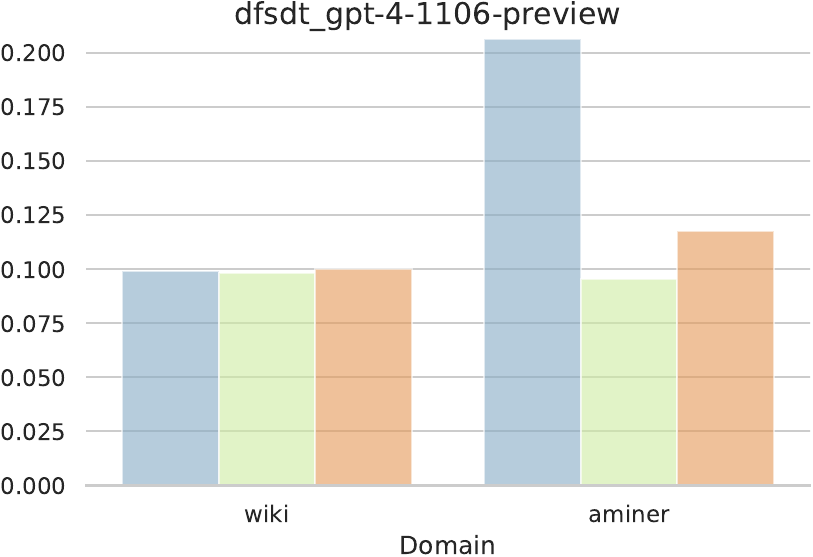}
    }
    \subfigure[FC]{
        \includegraphics[width=0.3\textwidth]{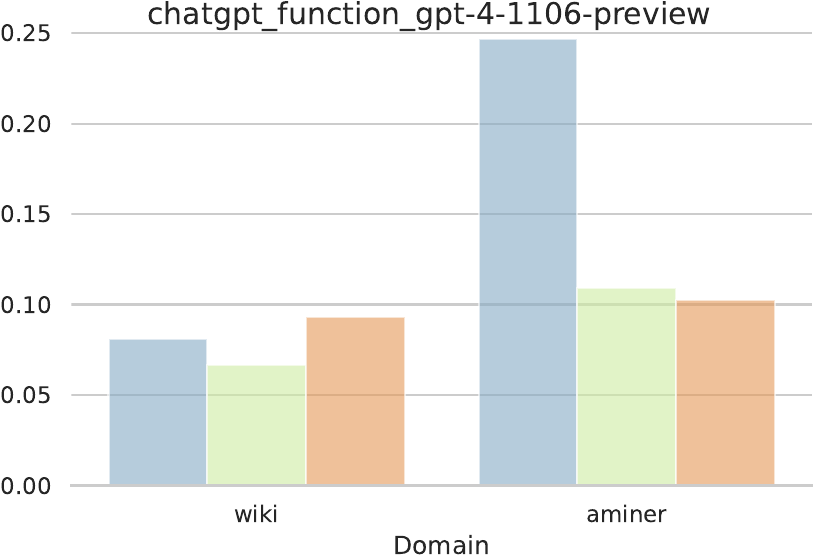}
    }
    \caption{Average performance of GPT-4 with different $3$ workflows on all tasks. }
    \label{fig:all_performance_appendix}
\end{figure*}

\section{Evaluation Details}
\label{sec:eval_detail}

In our study, we evaluate different RALLMs on various datasets and domains under a one-shot setting. On each task, we will give RALLMs an example of how to interact with the domain API to retrieve the knowledge and reasoning on it. For the fairness of comparison, different RALLMs share the same example (query and answer) while the retrieval and reasoning processes are customized for each RAG workflow. Note that we also provide an example pool with at least 1.8k cases for each task, as shown in Table ~\ref{tab:tasks}, where the user of our toolkit can adjust the number of used examples to fit their own few-shot settings.

The models participating in the evaluation fall into two categories: closed-source models that generate responses via API calls, and open-source models that are directly deployed for inference, with a temperature parameter set to $0$. The other hyper-parameters of open-source models are summarized in Table ~\ref{tab:hyper_parameters}.  To load open-source models, we employ the widely adopted \textit{PyTorch} and \textit{transformers} libraries. The evaluation experiments are executed on an Ubuntu $20.04.4$ server, furnished with $112$ Intel Xeon(R) Platinum $8336$C CPU cores, and complemented by graphic cards incorporating $8$ NVIDIA A100 SXM $80$GB GPUs.  Furthermore, the software environment includes CUDA version $11.4$, Python version $3.9.17$, PyTorch version $2.1.2$, and the transformers library version $4.28.1$.

\begin{figure*}[ht]
    \centering
    \subfigure[ReAct]{
        \includegraphics[width=0.23\textwidth]{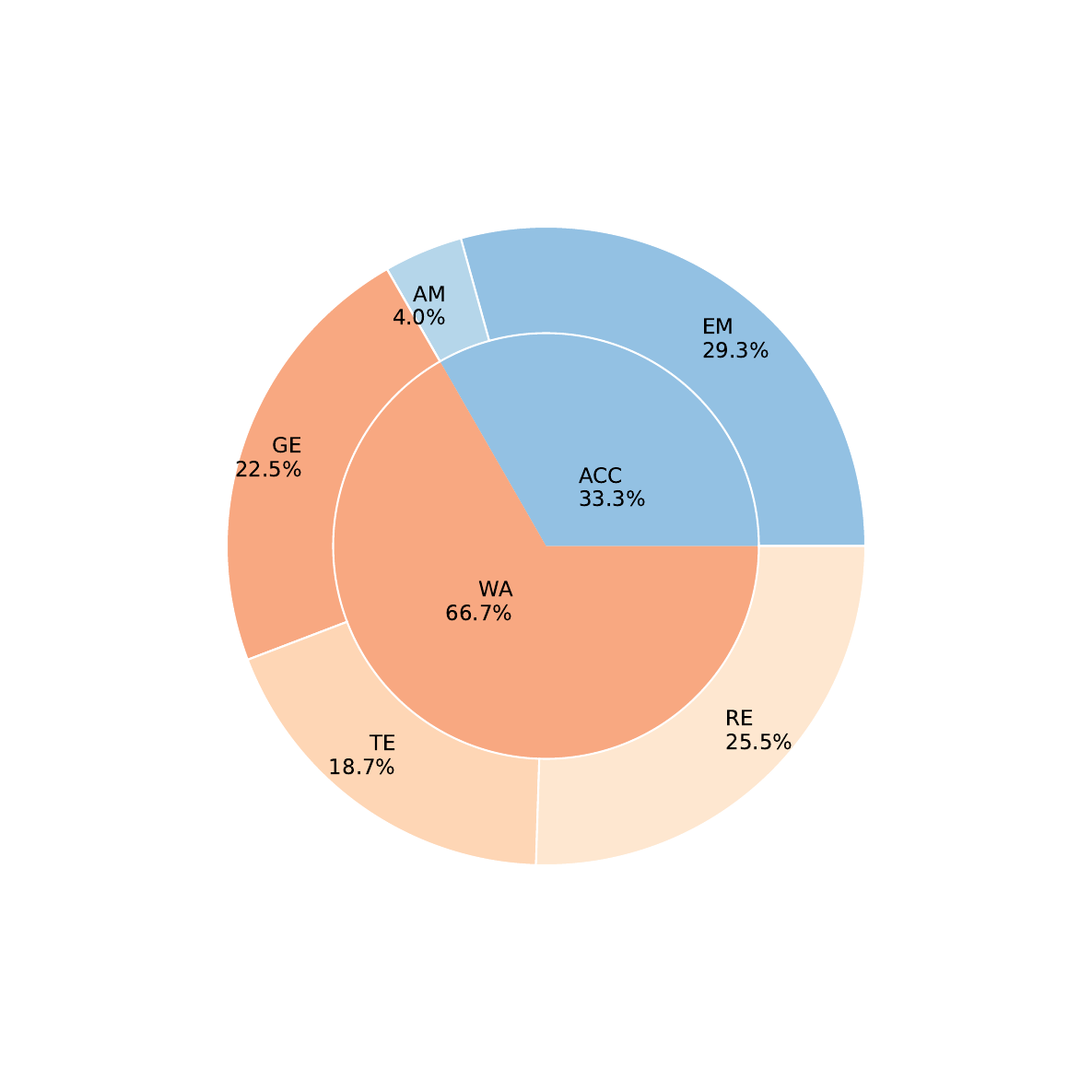}% 设置每张图的宽度为总宽度的四分之一
    }
    \subfigure[PAL]{
        \includegraphics[width=0.23\textwidth]{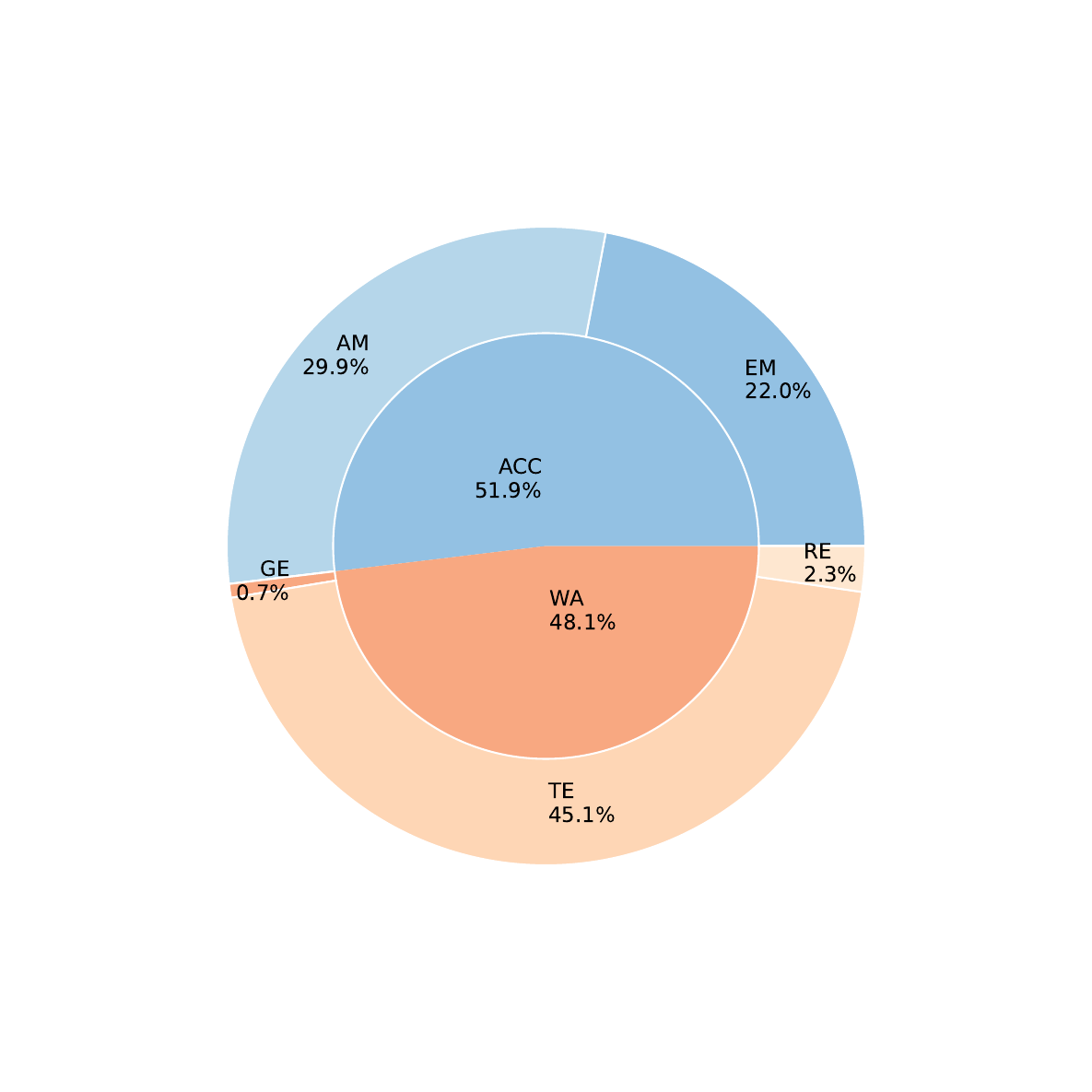}
    }
    \subfigure[DFSDT]{
        \includegraphics[width=0.23\textwidth]{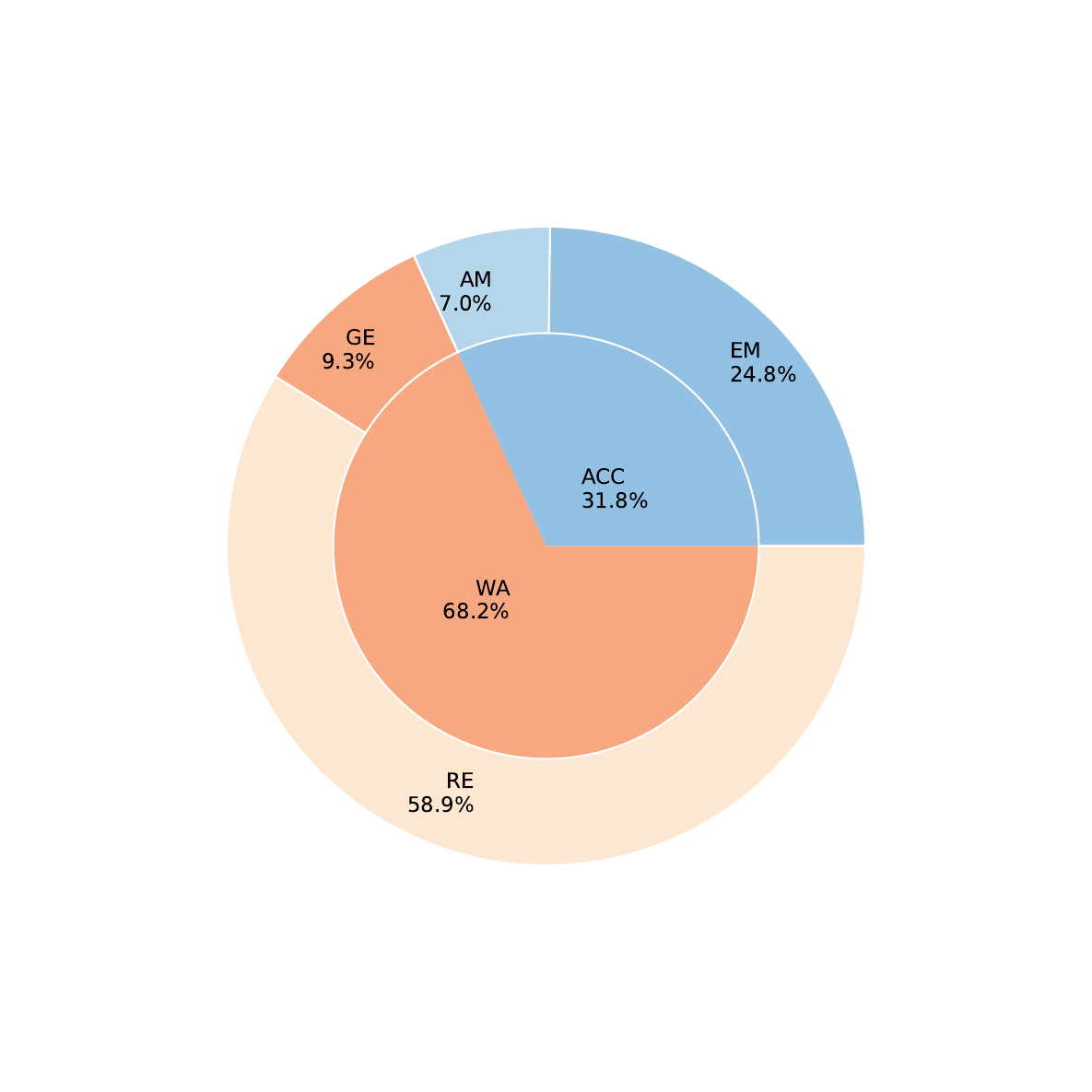}
    }
    \subfigure[FC]{
        \includegraphics[width=0.23\textwidth]{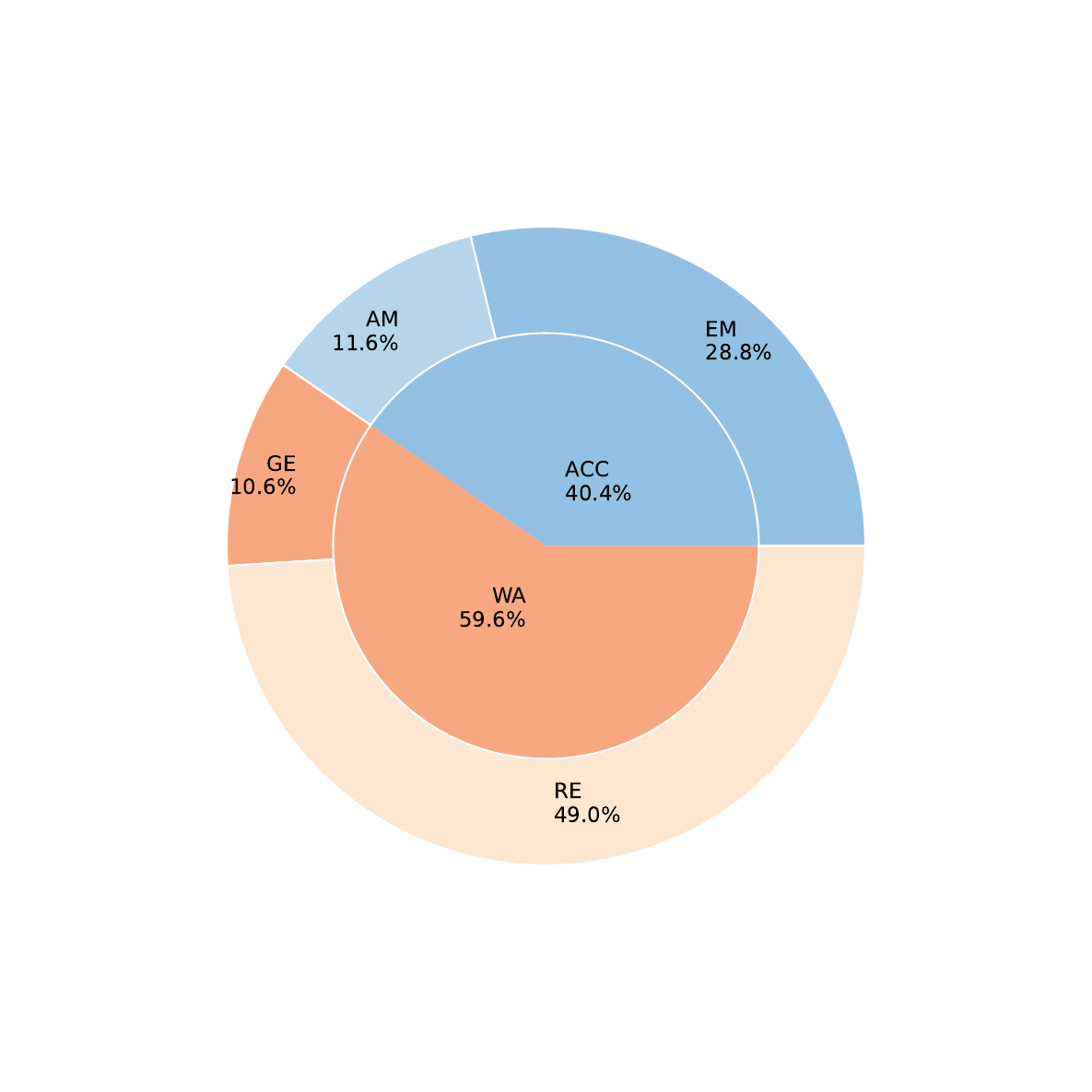}
    }
    \caption{Error distribution of GPT-3.5-turbo-1104 with different $4$ workflows on all tasks. }
    \label{fig:all_error_appendix}
\end{figure*}

Given that all tasks are arranged in open-ended generative question-answering formats, we choose F1 score as evaluation metrics to reflect the model's genuine performance. The metric is specifically tailored to the characteristics of different task levels and is used for post-processing the answers. For tasks that primarily use F1 as the evaluation metric, we employ a relaxed version of F1 (token match rate). Specifically, after tokenizing the model's predicted results and the reference answers using the GPT2Tokenizer~\citep{radford2019language}, we calculate whether each token in the prediction appears in the corresponding position of the gold standard.

\section{Extra Experiment Results}
\label{app:deploy}

\subsection{Wiki Domain Results}
\textbf{RQ7}: \textit{How effective are RALLMs on wiki  domain?}

The performance of different RALLMs on Wikipedia domain is demonstrated in Table~\ref{tab:comparison_wiki}. In the Wikipedia domain, which contains over 6.6 million English articles, some models with ReAct workflow, such as ReAct with GPT-4-1106, demonstrate strong performance across all three levels of tasks and win the first place on all these tasks. To illustrate the performance difference among RAG workflows, we display the average performance of GPT-4-1106 model with ReAct, DFSDT and FC  workflow in Figure~\ref{fig:all_performance_appendix}. The DFSDT and FC perform badly on the three levels' tasks of Wikipedia domain, where their F1 scores are all below 0.1, which is much lower than the ReAct workflow's scores even though they are all based on GPT-4-1106. This suggests that the ReAct workflow is effective at retrieving, understanding, and applying knowledge in a broad open-domain context.

\subsection{Error Analysis for GPT-3.5-turbo}

\textbf{RQ8}: \textit{What kinds of errors does gpt-3.5 make on different workflows?}

% Previously in Section ~\ref{analysis_tools}, we have defined the error types of RALLM systems. In this section, we attempt to identify the proportion of gpt-4's error types on different RAG workflows. 
As shown in Figure~\ref{fig:all_error_appendix}, the error distribution of gpt-3.5-turbo is different from gpt-4 in Figure~\ref{fig:all_error}. We find that the PAL workflow with gpt-3.5-turbo has a larger Answer Match and EM portion than the PAL workflow with gpt-4.  While their AM portion are similar (29.9\% vs. 27.1\%), gpt-3.5-turbo has a  significantly larger EM portion (22.0\%) than gpt-4 (16.5\%). However, for other workflows (ReAct, DFSDT, FC), gpt-3.5-turbo perfoms much worse than gpt-4. The main error gpt-3.5-turbo encountered is RE (reasoning error), especially on DFSDT and FC. These results reflect that gpt-3.5-turbo may be more dependant on the retrieved knowledge for reasoning while gpt-4 are better at inner knowledge. This provides a crucial insight about the  difference between gpt-3.5-turbo and gpt-4 on their response and error types with different workflows.

\begin{figure}[t]
    \centering
        \includegraphics[width=0.5\textwidth]{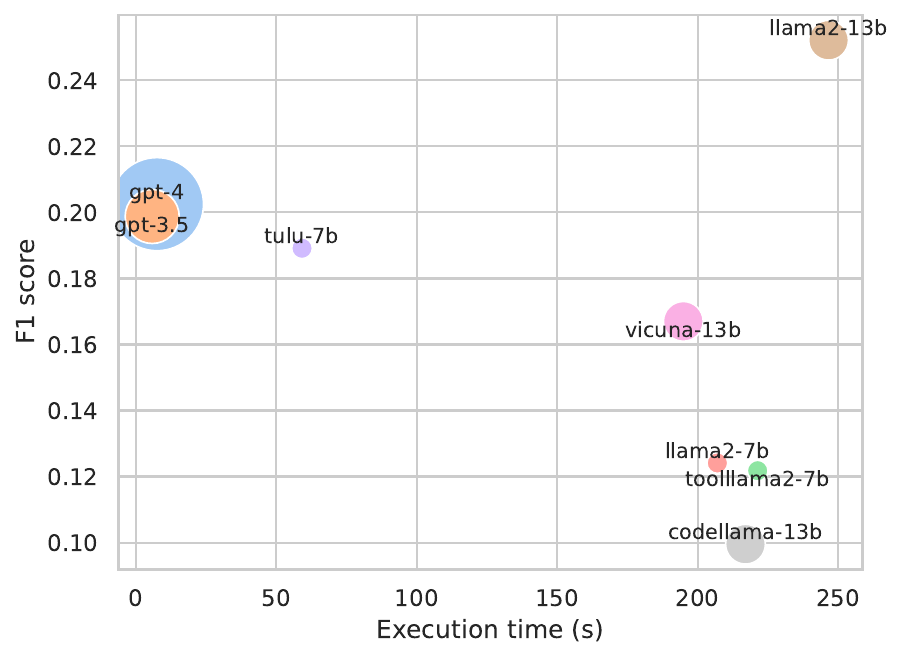}
    \caption{Average time cost and task performance of different LLMs with PAL workflow on wiki domain for deployment.  }
    \label{fig:deploy_wiki}
\end{figure}

\subsection{Deployment Analysis for Wiki Domain}

\textbf{RQ9}: \textit{Which system provides the best practical performance (efficiency and effectiveness) on wiki  domain?}

We are also interested in evaluating the practical performance of these RALLM systems on Wikipedia domain in terms of execution time and answer F1 score. We report the results of 8 LLMs with PAL workflow in Figure~\ref{fig:deploy_wiki}, since PAL has the simple-turn interaction with the environment, making it much faster than other workflow. Surprisingly, we find that although GPT-4 and GPT-3.5 get better F1 scores than other models on Aminer, they are not the best model in the Wikipedia domain.

If we only compare open-source LLMs, we find that although llama-13b is the best on F1 score, which is even higher than GPT-4, it is much slower other LLMs (250s per query on the Wikipedia domain). Among all the models, tulu-7b still achieves a good trade-off between efficiency (50s per query) and effectiveness (0.19 F1 score) on the Wikipedia domain. %Interestingly from Figure ~\ref{fig:deploy_wiki}, we find the open-source LLMs that based on Llama2 are generally  worse on the efficiency than those Llama-based LLMs. For example, vicuna-13b is fine-tuned on Llama while Llama2-7b is in the Llama2 family and has less parameters than vicuna-13b. But, llama2-7b is slower than vicuna-13b. This finding can help developers choosing the suitable LLM for their domain applications.

\end{document}